\documentclass[runningheads]{llncs}

\usepackage{multibib}

\usepackage{graphicx}

\usepackage{tikz}
\usepackage{comment}
\usepackage{amsmath,amssymb} 
\usepackage{color}
\usepackage{xcolor}
\definecolor{shadecolor}{RGB}{220,220,220}

\usepackage{makecell}
\usepackage{multirow}
\usepackage{orcidlink}
\usepackage{breakcites}
\usepackage[accsupp]{axessibility}

\usepackage{times}
\usepackage{epsfig}

\usepackage{mathalias}
\usepackage{booktabs}
\usepackage{verbatim}
\usepackage{enumitem}
\usepackage{subfig}
\usepackage{mwe}
\usepackage{wrapfig}

\usepackage[normalem]{ulem}

\newcommand{\etal}{\textit{et al}. }
\newcommand{\ie}{\textit{i}.\textit{e}., }
\newcommand{\eg}{\textit{e}.\textit{g}., }

\newcommand{\mybox}[1]{\par\noindent\colorbox{shadecolor}
{\fbox{\parbox{\dimexpr\textwidth-4\fboxsep\relax}{#1}}}}

\usepackage[accsupp]{axessibility}  

\newcommand{\new}[1]{#1}
\DeclareMathOperator{\Loss}{\mathcal{L}}

\usepackage[export]{adjustbox}

\newcommand{\adjimage}[1]{\adjustbox{valign=m,vspace=1pt}{\includegraphics[width=.2\linewidth]{#1}}}

\newcommand{\imgid}{01591}
\newcommand{\imgage}{20}

\begin{document}

\pagestyle{headings}
\mainmatter
\def\ECCVSubNumber{4880}  

\title{Custom Structure Preservation in Face Aging} 

\titlerunning{Custom Structure Preservation in Face Aging}

\author{Guillermo Gomez-Trenado\inst{1}
\orcidlink{0000-0003-3366-6047} 
\and
Stéphane Lathuilière\inst{2}
\orcidlink{0000-0001-6927-8930} 
\and
Pablo Mesejo\inst{1} \and
Óscar Cordón\inst{1}}
\authorrunning{Gomez-Trenado et al.}

\institute{
DaSCI research institute, DECSAI, University of Granada, Granada, Spain \email{\{guillermogomez,pmesejo,ocordon\}@ugr.es}
\and
LTCI, Télécom-Paris, Intitute Polytechnique de Paris, Palaiseau, France
\email{stephane.lathuiliere@telecom-paris.fr}}

\maketitle

\begin{abstract}
In this work, we propose a novel architecture for face age editing that can produce structural modifications while maintaining relevant details present in the original image. We disentangle the style and content of the input image and propose a new decoder network that adopts a style-based strategy to combine the style and content representations of the input image while conditioning the output on the target age. We go beyond existing aging methods allowing users to adjust the degree of structure preservation in the input image during inference. To this purpose, we introduce a masking mechanism, the CUstom Structure Preservation module, that distinguishes relevant regions in the input image from those that should be discarded. CUSP requires no additional supervision. Finally, our quantitative and qualitative analysis which include a user study, show that our method outperforms prior art and demonstrates the effectiveness of our strategy regarding image editing and adjustable structure preservation.
Code and pretrained models are available at \url{https://github.com/guillermogotre/CUSP}.
\end{abstract}

\section{Introduction}

Face age editing \cite{fu2010age,kemelmacher2014illumination,wang2016recurrent}, or aging, consists in automatically modifying an input face image to alter the age of the depicted person while preserving identity. Over the last few years, this problem has attracted a growing interest because of its numerous applications. In particular, it is used in the movie production industry to edit actors' faces or in forensic facial approximation to reconstruct the faces of missing people. The advances in deep learning methods unlock the development of fully automatic edition algorithms that avoid hours of makeup and post-production retouching.

\begin{figure}[t]
    \centering
    \includegraphics[width=0.7\textwidth]{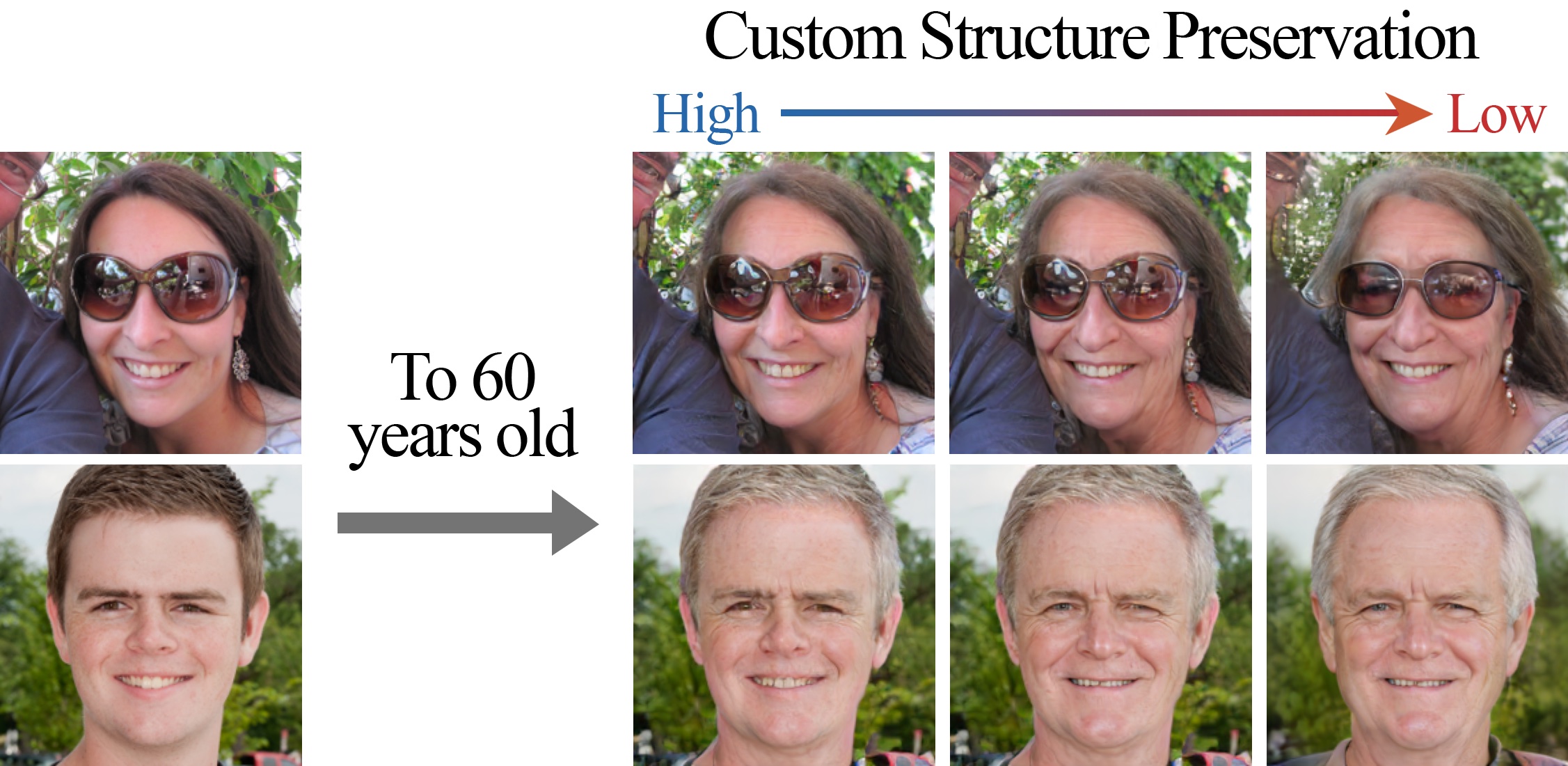}
    \caption{The user can choose the degree of structure preservation at inference time. Facial morphology transformations are more profound as we move to the right (lower structure preservation). }
    \label{fig:teaser}
\end{figure}%

Recent deep learning approaches adopt an encoder-decoder architecture ~\cite{antipov2017face,He2019S2GANSA,reaging2021,orel2020lifespan,wang2016recurrent,wang2018face,hrfae2021,zhang2017age}. The image is encoded in a latent space that can be modified depending on the target age and fed to a decoder that generates the output image. The overall network is usually trained using a combination of losses that assess image quality, identity preservation, and age matching. However, despite the success of all these approaches, face editing remains challenging, and current methods usually fail when faced with sizeable differences between the age of the person displayed in the input image and the target age. Indeed, most approaches~\cite{antipov2017face,He2019S2GANSA,wang2016recurrent,wang2018face,hrfae2021,zhang2017age} only superficially modify the skin's texture while the face's shape is kept unchanged. These approaches fail with significant age gaps since face shape can change significantly during a lifetime. Few methods try to go beyond some limited age gaps, but they either consider only a tightly cropped face region~\cite{kemelmacher2014illumination,wang2016recurrent} or require specific pre-processing involving an image segmentation step~\cite{orel2020lifespan}.

This work proposes a novel framework that allows profound structural changes in facial transformations. This framework achieves a realistic image transformation with age gaps that imply changes in head shape or hair growth. In addition, we argue that the face editing task is an ill-posed problem because every person gets older in a different and non-deterministic way: some people drastically change, while others are easily recognizable in old photographs. In this sense, we propose a methodology that allows the user to adjust, at inference time, the degree of structure preservation. Thus, the user can provide an image and obtain different transformations where the structure (\ie face shape or hair growth) is preserved at different levels. Fig. \ref{fig:teaser} shows some qualitative results obtained with our method. 
Furthermore, the user can choose different degrees of structure preservation: with high preservation, the model only changes the texture, while with lower preservation, the shape of the face is also modified.

\noindent The contributions of this paper can be summarized as follows:
\begin{itemize}[noitemsep,topsep=0pt]
\item We propose a novel architecture for face age editing that can produce structural modifications in the input image while maintaining relevant details present in the original image. We take advantage of recent advances in image-to-image (I2I) translation \cite{huang2018multimodal,lee2018diverse} and unconditional image generation \cite{Karras2019stylegan2} to design our architecture. We disentangle the style and content of the input image, and we propose a new decoder network that adopts a style-based strategy to combine the style and content representations of the input image while conditioning the output on the target age.
    \item We go beyond existing aging methods allowing the user to adjust the degree of structure preservation in the input image at inference time. To this aim, we introduce a masking mechanism, through a so-called CUstom Structure Preservation (CUSP) module, that identifies the relevant regions in the input image that should be preserved and those where details are irrelevant to the task. Importantly, our mechanism for adjustable structural preservation does not require additional training supervision.
    \item Experimentally, we show that our method outperforms existing approaches in three publicly available high-resolution datasets and demonstrate the effectiveness of our mechanism for adjusting structure preservation.
\end{itemize}

\section{Related Work}
\label{sec:related}

Most recent approaches for \textbf{face aging} adopt a similar strategy based on an encoder-decoder architecture~\cite{antipov2017face,He2019S2GANSA,reaging2021,orel2020lifespan,wang2016recurrent,wang2018face,hrfae2021,zhang2017age}. In these methods, the input image is projected onto a latent space where content is manipulated before decoding the output image. Some methods~\cite{wang2018face,alaluf2021matter} add an identity term to the total loss to better ensure the preservation of the identity during the translation process. These methods principally differ in the choice of the network architecture and the manner the latent representation is manipulated. For instance, Wang \etal~\cite{wang2016recurrent} introduce a recurrent neural network to iteratively alter the image, while in \cite{hrfae2021}, the latent image representation is modified using a simple affine transformation. Re-AgingGAN~\cite{reaging2021} employs an age modulator that outputs transformations that are applied then to the decoder, and Or-El \etal \cite{orel2020lifespan} adopt a multi-domain translation formulation, showing that segmentation information can be leveraged to improve aging. In our work, we adopt an encoder-decoder framework similar to~\cite{He2019S2GANSA,hrfae2021}. However, our approach goes beyond existing methods that generate a single image for a given image-target age pair. Indeed, we offer the user the possibility to adjust the degree of structure preservation during translation, and, in this way, we can output a set of plausible resulting facial images.

Our method also leverages recent advances from the \textbf{I2I translation} research area. I2I translation consists in learning a mapping between two visual domains. In the pioneering work of Isola \etal \cite{isola2017image}, an encoder-decoder network is trained using a dataset composed of image pairs from the two domains. Later, many works addressed I2I translation in an unpaired setting, assuming two independent sets of images of each domain~\cite{fu2019geometry,Liu_nips2017,zhu2017unpaired}. 
These works, of which cycleGAN~\cite{zhu2017unpaired} is a paradigmatic example, mainly focus on introducing regularization mechanisms when training the I2I translation models. 
 Another research direction is designing more advanced architectures to improve image quality or obtain several possible outputs for a given input \cite{huang2018multimodal,lee2018diverse,zhu2017multimodal}. 
 \new{Disentangling style and content information has led to both higher image quality and diversity \cite{huang2018multimodal,park2020swapping}.}
 We adopt a similar strategy in order to allow custom structure preservation. Thanks to this strategy, our CUPS module can act on the spatial information passing through the content branch while preserving style information.


\textbf{Style-based architectures} recently attracted much attention for the problem of unconditional image generation. In particular, StyleGAN2~\cite{Karras2019stylegan2} is now used in many face manipulation tasks~\cite{richardson21cvpr,Yao_2021_ICCV}.  \new{In the case of face aging, \cite{alaluf2021matter} uses a pretrained StyleGAN2 model~\cite{Karras2019stylegan2} equipped with a pSp encoder~\cite{richardson21cvpr}, and an age classifier~\cite{rothe2018dex} to tailor an age editing model with unlabeled data.} In StyleGAN2, a network maps a Gaussian latent space onto style vectors; these vectors are later combined via a convolutional network to produce the output image. Finally, the synthesis network aggregates the style vectors through modulation operations. We take inspiration from the StyleGAN2 generator to design a novel decoder that combines the input style and the target age with the content representation via weight demodulation.

Regarding the more general \textbf{image editing} problem, our method shares similarities with several approaches employing masking mechanisms or attention maps  to preserve relevant parts in the input image \cite{ak2019attribute,Kim_2021_CVPR,pumarola2018ganimation,tang2019attention}.
 For instance, mask consistency is employed in~\cite{Kim_2021_CVPR} to improve multi-domain translations. As in our approach, masks are estimated using the guided backpropagation (GB) algorithm~\cite{guidedbackprop2015}. In the case of facial images, a mask  is employed in GANimation~\cite{pumarola2018ganimation} to different regions that should be preserved and those that should be modified to change the facial expression. In GANimation, masks are predicted by the main network, while we employ an auxiliary network and GB~\cite{guidedbackprop2015} to obtain the mask.

\section{Proposed Method}
In this work, we address the face age editing problem. Therefore, our goal is to train a network able to transform an input image $\Xmat$, such that the person depicted looks like being of the target age $a_t$.  At training time, we assume that we have at our disposal a dataset composed of $I$ face images of resolution $H\times W$, such that $\Xmat_i\!\in\!\mathbb{R}^{H\times W\times3}, i=1,...,I$  with their corresponding age label $a_i\in \{1,..N\}$. Note that the age labels are automatically obtained using a pre-trained age classifier. Similar to previous approaches~\cite{hrfae2021,zhang2017age}, we employ the DEX classifier~\cite{dex2015}.

One of the main difficulties lies in modifying the relevant details in the input image while preserving non-age-related regions. To this aim, we introduce a style-based architecture detailed in Sec.\ref{sec:archi}. In contrast to previous works, the CUSP module allows the user to indicate the desired level of structure preservation through two parameters: $\sigma_m>0$ and $\sigma_g>0$. These parameters act locally and globally, respectively, as detailed later. The proposed CUSP module is described in Sec.~\ref{sec:cusp}. Finally, we present the whole training procedure in Sec.~\ref{sec:training}.

\begin{figure}[t]
    \centering
        \includegraphics[width=1\textwidth]{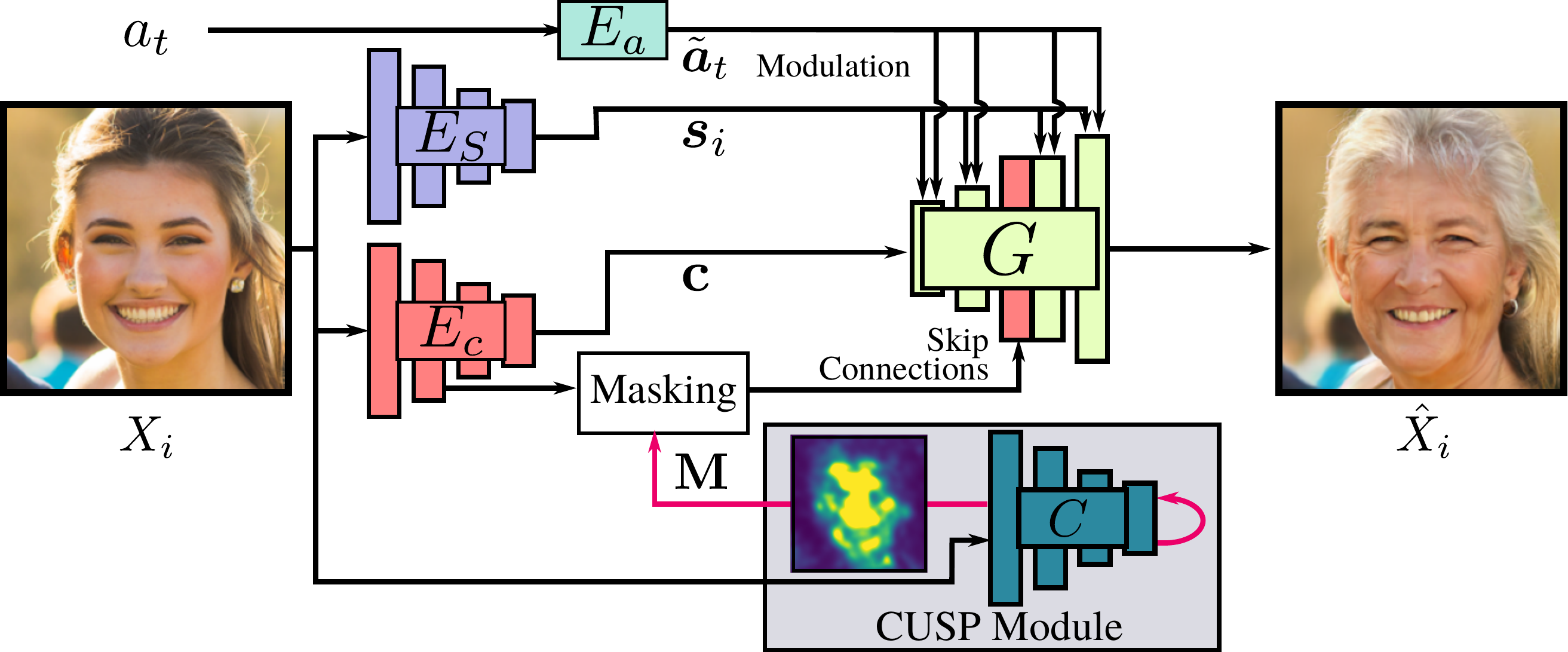}
        \caption{Illustration of the proposed approach. A style encoder $E_s$ extracts a style representation of the input image $\Xmat_i$. A content encoder $E_c$ encodes spatial information. Target age $a_t$ is embedded using a multi-layer perceptron $E_a$. Our generator $G$ outputs the image $\hat{\Xmat}_i$ by combining the input style and content representations conditioned on the target age. Our CUSP module predicts a blurring mask $\Mmat$ applied to the skip connections to allow the user to choose a CUstom level of Structure Preservation.}
        \label{fig_diag}
\end{figure}

\subsection{Style-based Encoder-decoder}
\label{sec:archi}
As illustrated in Fig. \ref{fig_diag}, our architecture employs five different networks:
(1) A style encoder $E_s$ extracts a style representation $\svect_i$ of the input image $\Xmat_i$. $E_s$ discards any spatial information via global-average-pooling at the last layer. The use of a style encoder allows global information to be used at any location in the decoder.
(2) A content encoder $E_c$ outputs a tensor $\cmat$ describing the content of the input image. Contrary to $E_s$, the content encoder preserves spatial and local information. In our case, the use of separated style and content encoders is justified by the fact that our CUSP module should not affect the image style $\svect_i$ but only the structure of the image. (3) An 8-layers fully connected network, $E_a$, embeds the target age $a_t$: $\tilde{\avect}_t=E_a(a_t)$.  (4) An image generator $G$ estimates the output image $\hat{\Xmat}_i$ by combining the style and content representations with the target age embedding $\tilde{\avect}_t$. (5) Finally, our CUSP module allows the user to choose the level of structure preservation. This module predicts a mask $\Mmat$ used to act on the skip connections between the content encoder and the decoder. More precisely, we blur the regions indicated by the mask $\Mmat$ to propagate only the non-age-related structural information to the decoder.

\label{sec:gen}
Our image generator $G$ is designed to combine the outputs of the style and content encoders with the target age embedding. Its architecture is inspired by StyleGAN2~\cite{stylegan2}, which achieves state-of-the-art performance in unconditional image generation. However, we provide several modifications to tailor the architecture to the aging task. $G$ comprises a sequence of elementary blocks (see \emph{Supplementary Materials} for illustration). Differently from StyleGAN2, each block takes three inputs: the former block output feature map, the style encoding $\svect_i$, and the class embedding $\tilde{\avect}_t$. Each block is composed of two sub-blocks. In the first one, we use the style vector $\svect_t$ to modulate the convolution operations as in ~\cite{stylegan2}. In the second one, the age embedding is used for modulation. Up-sampling is applied to the input of the first sub-block. Similarly to \cite{stylegan2}, random noise is summed to the feature maps between each sub-block, while scaling and bias parameters (\ie $w$ and $b$) are learned for each sub-block. 

Note that all blocks are combined following the \textit{input skips} architecture of  StyleGAN2, where  a layer named \emph{tRGB} is introduced. 
Such layer predicts intermediate images at every resolution scaled and added to generate the final image. tRGB is also conditioned on the age embedding. A single skip connection is introduced before the last block, contrarily to U-Net \cite{unet2015} that includes them in every layer.

\subsection{CUSP Module}
\label{sec:cusp}

Skip connections (SC) \cite{unet2015} are efficient tools to provide high-frequency information to the decoder allowing accurate reconstruction~\cite{isola2017image}. High frequencies carry accurate spatial information that favors pixel-to-pixel alignment between inputs and outputs, as, for instance, needed in segmentation. However, previous works~\cite{siarohin2018deformable} show they are not suited for tasks where the input and output images are not pixel-to-pixel aligned. 
For example, input and output images are aligned when the age gap is small in the aging task. However, this assumption does not hold in every image region with significant gaps. This misalignment is particularly predominant in areas other than the background since facial morphology or hairstyle may change.

Therefore, we propose to control the amount of structural information that flows through the SC. This control is obtained by blurring the feature maps going through them. Nevertheless, every region should not be treated in the same way. For instance, depending on the task, the user may prefer to preserve the background while blurring the foreground to loosen conditioning on the input image in this region. Therefore, we propose a specific mechanism to identify relevant image regions for the translation.

\noindent\textbf{Mask Estimation.}
We employ an additional classification network $C$, pretrained to recognize the age of the person depicted on an image. We use the DEX classifier~\cite{dex2015} again. Since DEX is pretrained on 224$\times$224, the input image is rescaled to this resolution. Then, we apply the GB algorithm \cite{guidedbackprop2015} to obtain a
tensor $\Bmat\in \mathbb{R}^{224\times 224\times 3}$, where locations with higher norm correspond to regions predominantly used by the classifier. In other words, $\Bmat$ pinpoints relevant regions for the age classification task. GB points out the key areas to recognize the age and should, therefore, be modified by the aging network. Importantly, GB is usually used to visualize the regions that influence one specific network output (\ie one specific class)~\cite{guidedbackprop2015}. In our case, we apply GB to the sum of the classification layer before softmax normalization to obtain class-independent masks. We select GB \cite{guidedbackprop2015} over other approaches~\cite{gradcam2017,muhammad2020eigen,srinivas2019full} since it is a fast, simple, and strongly supported method for visualization.

We need to transform $\Bmat$ to obtain a mask $\Mmat\!\in\![0,1]^{224\times 224}$. We proceed in several steps. First, we average $\Bmat$ over the RGB channels, take the absolute value, and apply Gaussian blur to get smoother maps. In this way, we obtain a tensor  $\tilde{\Bmat}\in {\mathbb{R}_{>0}}^{224\times 224}$ that indicates relevant regions. To obtain values in $[0,1]$, we need to normalize $\tilde{\Bmat}$. Our preliminary experiments showed that after normalizing by twice the variance $\sigma$ of $\tilde{\Bmat}$ (over the locations),  
relevant areas for the aging task are close to 1 or above. We apply clipping to bring all those important regions to 1.
Formally the mask values are computed as follows:   
\begin{equation}
   \Mmat = \min \left( \frac{\tilde{\Bmat}}{2 \times \sigma},1 \right) 
\label{eq_cusp_mask}
\end{equation}
where \emph{min} denotes the element-wise minimum. Next, we detail how this mask is employed in our encoder-decoder architecture.

\noindent\textbf{Skip connection blurring.}
Assuming a feature map $\Fmat_c\!\in\!\mathbb{R}^{H'\times W'\times C}$ provided by the content encoder $E_c$, we resize $\Mmat$ to the dimension of $\Fmat_c$ obtaining a mask $\tilde{\Mmat}\!\in\![0,1]^{H'\times W'}$. We then blur $\Fmat_c$ using two different Gaussian kernels with variance $\sigma_m>0$ and $\sigma_g>0$. The variance $\sigma_m$ is applied in the region indicated by $\Mmat$, while $\sigma_g$ is used over the whole feature map. The motivation for this choice is that the user can choose to alter structure preservation locally, globally, or both. At training time, $\sigma_m$ and $\sigma_g$ are sampled randomly to force the generator $G$ to perform well for any blur parameter. At test time, both values might be provided by the user.  Formally, the blurred feature map is computed as follows:
\begin{equation}
    \tilde{\Fmat}_c = \;  \tilde{\Mmat} \circ (\Fmat_c * \kmat_m)  \; +   (1- \tilde{\Mmat}) \circ (\Fmat_c * \kmat_g)    
\label{eq_cusp}
\end{equation}
where $*$ denotes the convolution operation, $\circ$ is  the Hadamard product, and $\kmat_m$ and $\kmat_g$ are the Gaussian kernels of variances $\sigma_m$ and $\sigma_g$. 

\subsection{Overall Training Procedure}
\label{sec:training}

Training facial age editing models is particularly challenging since paired images are unavailable. Therefore, similarly to \cite{reaging2021,orel2020lifespan,hrfae2021}, our training strategy is either focused on reconstruction (when the target age matches the input age) or I2I translation (when the target age is different). Also, similar to \cite{reaging2021,orel2020lifespan,hrfae2021}, training is performed using a set of complementary losses described below.

\noindent\textbf{Reconstruction loss ($\Loss_r$).} When the target age $a_t$ is equal to the image age $a_i$, we expect to reconstruct the input image. We, therefore, adopt an L1 reconstruction loss: 
\begin{equation}
    \Loss_r = \|T(\Xmat_i,a_i) -\Xmat_i \|_1
\end{equation}
where $T$ denotes the whole aging network, which output is the scaled addition of every \emph{tRGB} block.

\noindent\textbf{Age fidelity losses ($\Loss_D, \Loss_C$).} 
Following \cite{choi2018stargan}, we use a conditional discriminator $D$ to asses that generated images correspond to the target age $a_t$. More precisely, we employ the discriminator architecture of StyleGAN2 equipped with a multiclass prediction head, together with the training loss $\Loss_D$ defined in ~\cite{miyato2018cgans}.

We employ a loss $\Loss_C$  that assesses age matching using the same pretrained classifier $C$ used in the CUSP module to complement the adversarial loss. Furthermore,
$\Loss_C$ is implemented using the Mean-Variance loss~\cite{pan2018mean}, a classification loss tailored for age estimation.

\noindent\textbf{Cycle-Consistency loss ($\Loss_{cy}$).}
Following \cite{zhu2017unpaired}, we adopt a cycle consistency $\Loss_{cy}$ to force the network to preserve details that are not specific to the age (\eg, background or face identity). $\Loss_{cy}$ is given by:
\begin{equation}
    \Loss_{cy} = \| \Xmat_i - T(T(\Xmat_i,a_t),a_i) \|_1
\end{equation}

\noindent\textbf{Full objective.} Finally, the total cost function can be written
\begin{equation}
    \min_M \max_D \lambda_r\Loss_r + \lambda_C\Loss_C + \lambda_D\Loss_D + \lambda_{cy}\Loss_{cy}
\label{eq:final}
\end{equation}
where $\lambda_r, \lambda_C, \lambda_D$, and $\lambda_{cy}$ are constant weights. 

\section{Experiments}

\subsection{Evaluation protocol and implementation}

Every paper employs different metrics, datasets, and tasks in the aging literature. Therefore, we include a large set of metrics, datasets, and tasks in our experiments to allow comparison with most existing methods. 

\noindent\textbf{Datasets.}
In this paper, we employ three widely-used, publicly available high-resolution datasets for face aging and analysis:\\
\noindent$\bullet$ \textit{FFHQ-RR}: Initially proposed in \cite{hrfae2021}, this aging dataset based on FFHQ \cite{stylegan1} comprises of 48K images depicting people from 20 to 69 years old. Because of this \emph{Restricted age Range}, we refer to this dataset as \emph{FFHQ-RR}. Images are downsampled to 224$\times$224.\\ 
\noindent$\bullet$  \textit{FFHQ-LS}: Tis aging dataset, introduced in \cite{orel2020lifespan}, is composed of the 70K images from FFHQ \cite{stylegan1}, manually labeled in 10 age clusters that try to capture both geometric and appearance changes throughout a person's life: 0-2, 3-6, 7-9, 10-14, 15-19, 20-29, 30-39, 40-49, 50-69 and 70+ years old. Consequently, this dataset is referred to as \emph{FFHQ-LS} because of its \emph{LifeSpan} age range. The resolution of these images is 256$\times$256 pixels.\\ 
\noindent$\bullet$  \textit{CelebA-HQ} \cite{karras2017pggan,liu2015deep}: 
It consists of 30K images at 1024$\times$1024 resolution, which we downsample to 224$\times$224 pixels. The only age-related label in the dataset is \textit{young}, which can be either true or false.

The use of \emph{FFHQ-RR} and \emph{FFHQ-LS} may seem redundant since they are both based on the FFHQ dataset, but we perform distinct experiments on both datasets to allow comparison with existing state-of-the-art methods (which report results on at least one of them).


\noindent\textbf{Tasks.}
 We employ two tasks to evaluate the performance:\\
          \noindent $\bullet$ \textit{Young} $\rightarrow$ \textit{Old}: as in  \cite{hrfae2021}, we sample 1000 images belonging to the ``young'' category and translate them to a target age of 60. This task is only performed on CelebA-HQ.\\ 
         \noindent $\bullet$ \textit{Age group comparison}: similarly to \cite{reaging2021}, we consider different age groups: (20-29), (30-39), (40-49), and (50-69) on \textit{FFHQ-RR} and additionally (0-2), (3-6), (7,9), (15,19) on \textit{FFHQ-LS}. We again sample the first 1000 test images and translate every one of them into the central age of each of the four different age groups (25, 35, 45, and 55, respectively). 


\noindent\textbf{Metrics. }
We choose metrics to evaluate the two main  aspects of the aging task. Firstly, the translated/generated images must preserve the content of the input image in terms of identity, facial expression, and background. Secondly, the age translation might be accurate. 
In particular, we adopt the following metrics:\\
\noindent$\bullet$ \textit{LPIPS} \cite{lpips2018} measures the perceptual similarity when the target age coincides with the input image age.\\ 
\noindent$\bullet$ \textit{Age Mean Absolute Error (MAE)}. We employ a pretrained and independent age estimation network to compare the predicted age with the target age given an input image. As we already use the DEX pretrained classifier \cite{dex2015} at training time, we utilize Face++ API \footnote{Face++ Face detection API: \url{https://www.faceplusplus.com/} (last visited on \today).}. Experiments show that DEX is more biased towards younger age predictions than Face++. Therefore, reporting the MAE to the input target age $a_t$ would be biased. To compensate for this DEX-Face++ misalignment, we estimate the age of the original images with Face++ and compute the mean for each group. We then report the distance between the mean group predicted age and the transformed image predicted age. \\

\new{
\noindent$\bullet$ \textit{Kernel-Inception Distance \cite{binkowski2018kid} (KID)} assesses that the generated images are similar to real ones for similar ages. While FID \cite{heusel2017fid} is adopted in \cite{reaging2021}, we adopted KID as it is better suited for smaller datasets. We report the KID between original and generated images within the same age groups.\\
}
\noindent $\bullet$  \textit{Gender}, \textit{Smile}, and \textit{Face expression} preservation and \textit{Blurriness}: Face++ provides these metrics to evaluate input image preservation and quality. \textit{Gender}, \textit{Smile}, and \textit{Face expression} preservation are reported in percentages as in \cite{hrfae2021}.

\noindent\textbf{Implementation details.} 
We use the same training settings as StyleGAN2-ADA \cite{karras2020training} with $\lambda_r=10$, $\lambda_C=0.06$, $\lambda_D=1$, $\lambda_{cy}=10$. The optimizer used is Adam with $lr=0.0025$ and $\beta_1=0$, $\beta_2 =0.99$. 

FFHQ-RR and CelebA-HQ models are trained for 65 epochs with a batch size of 18. FFHQ-LS is trained for 140 epochs with a batch size of 16. All experiments are run on a single Nvidia A100 GPU. 

\begin{figure}[t]
\centering
\includegraphics[width=1\linewidth]{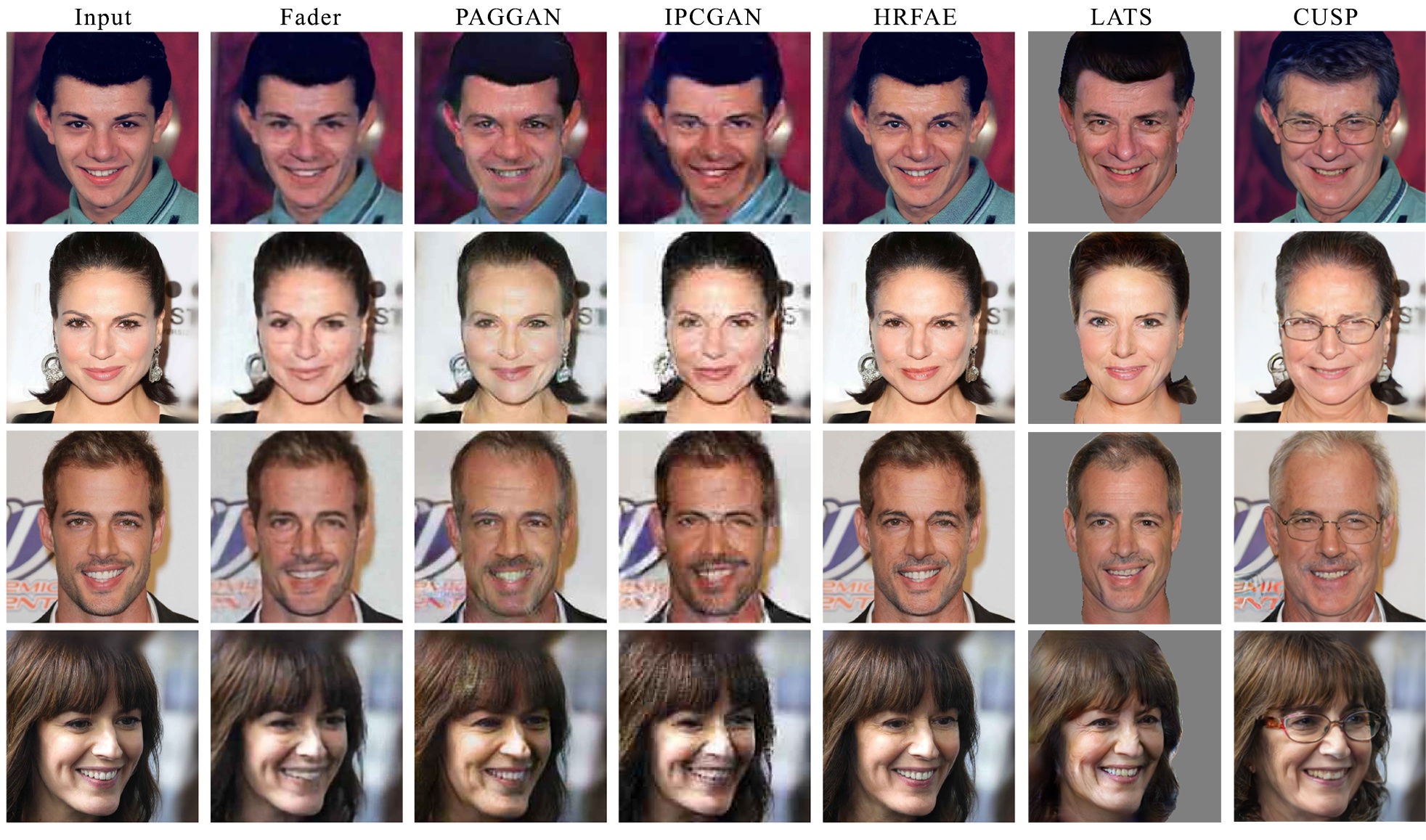}
 \caption{Comparison with State-of-the-art on CelebA-HQ for the \textit{Young} $\rightarrow$ \textit{Old} task employing a target age of 60 years old. }  \vspace{-0.3cm}
  \label{fig:hrfae_sota}
  
\end{figure}

\subsection{Comparison with State-of-the-Art}

From our literature review (Sec.~\ref{sec:related}), we identify HRFAE \cite{hrfae2021} and LATS \cite{orel2020lifespan} as the two main competing methods. Indeed, Re-aging GAN~\cite{reaging2021} cannot be included in the comparison since neither the code nor the age classifier used for evaluation is publicly available. Since HRFAE and LATS report experiments on different datasets and follow different protocols, we perform experiments using the two tasks previously described. First, we follow HRFAE \cite{hrfae2021}, which employs the \textit{Young} $\rightarrow$ \textit{Old} task on  \emph{CelebA-HQ}. In this case, the performance of FaderNet \cite{lample2017fader}, PAG-GAN \cite{yang2018learning}, IPC-GAN \cite{wang2018face}, and HRFAE (on 1024$\times$1024 resolution images) is reported in \cite{hrfae2021} and is included in our experimental comparison.  Second, we employ the \textit{age group comparison} task to allow better comparison with LATS \cite{orel2020lifespan} on the most challenging \emph{FFHQ-LS} dataset. Indeed, since no automatic quantitative evaluation is reported on the \emph{FFHQ-LS} in \cite{orel2020lifespan}, we chose the \textit{age group comparison} task that provides richer analysis than the \textit{Young} $\rightarrow$ \textit{Old} task.

\begin{figure}[t]
\begin{minipage}{0.43\linewidth}
\centering
\includegraphics[width=\columnwidth]{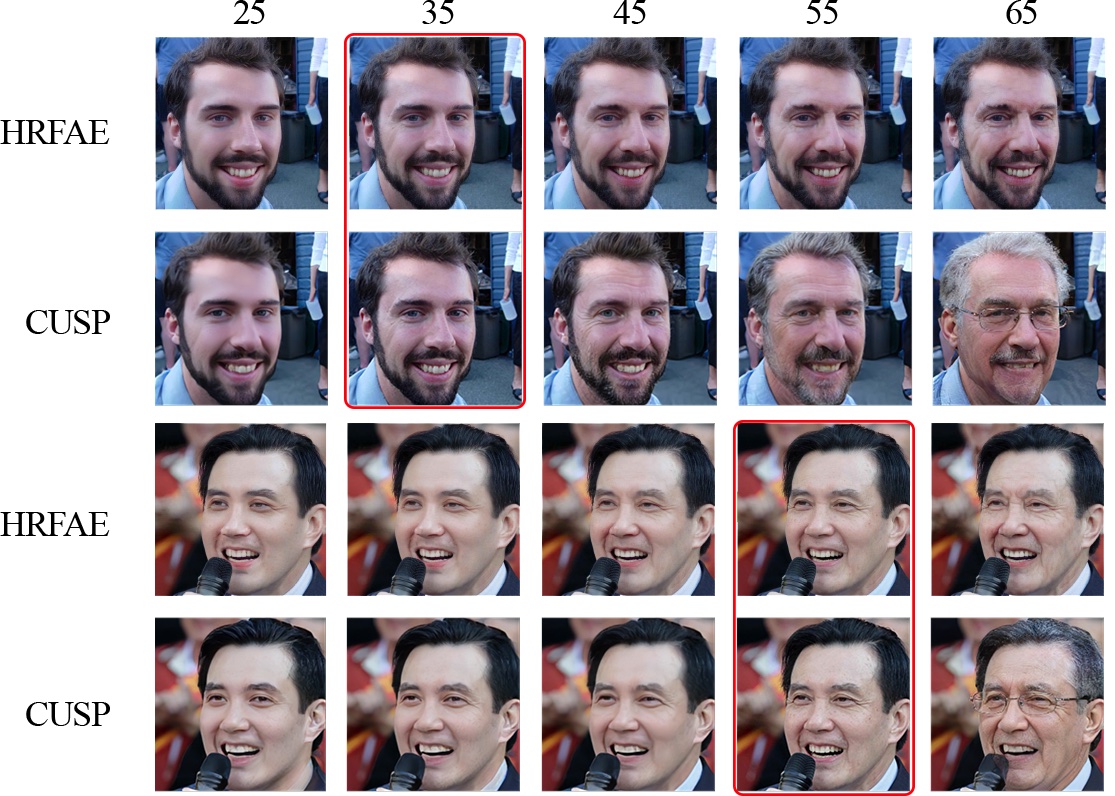}
  \caption{ Qualitative comparison with HRFAE. The images corresponding to the input ages are highlighted with red frames.}
  \label{fig:hrfae_comp}
\end{minipage}
\hfill
\begin{minipage}{0.53\linewidth}
\includegraphics[width=\columnwidth]{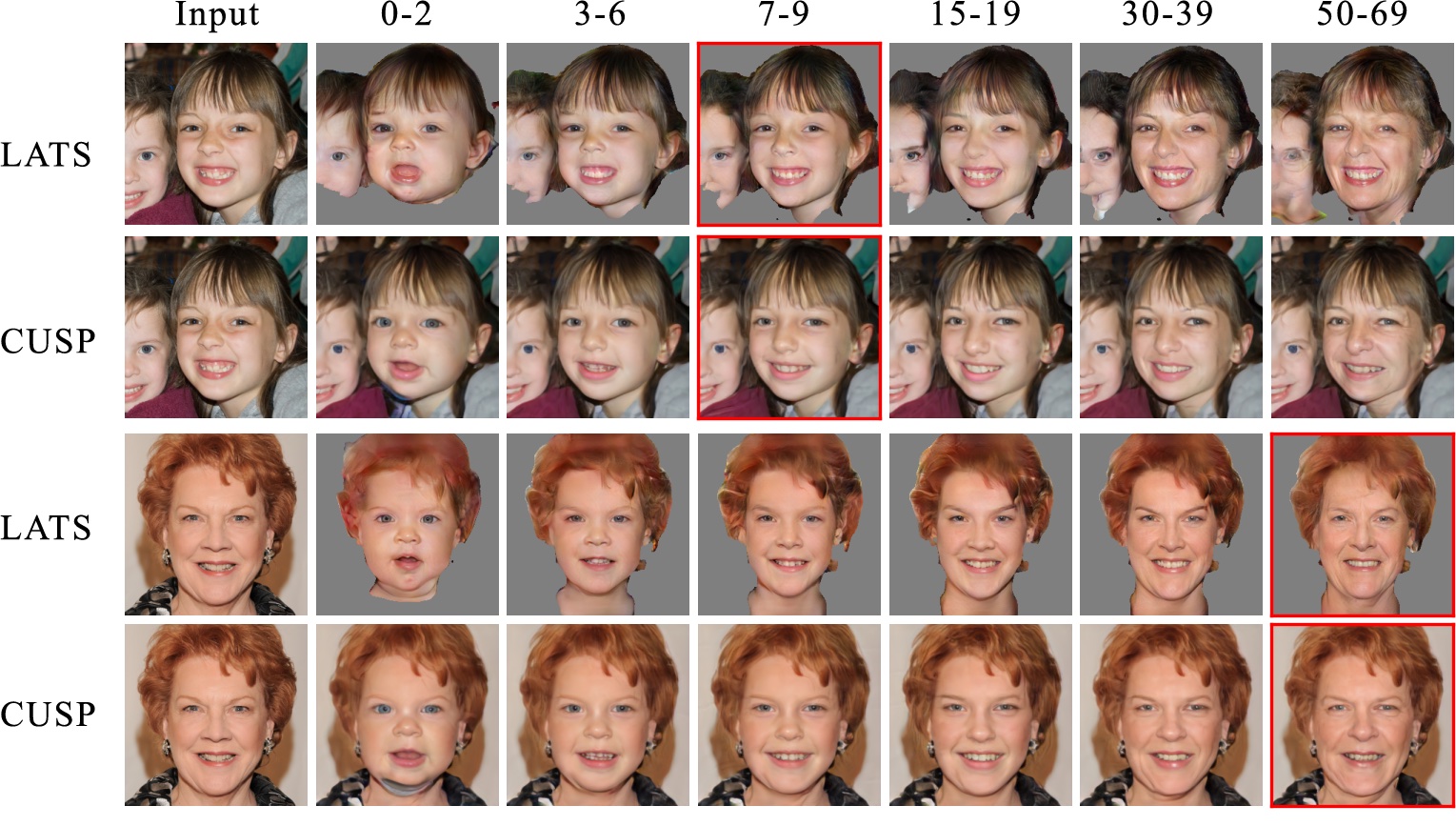}
  \caption{LATS comparison for different age targets. The images corresponding to the input ages are highlighted with red frames.}
  \label{fig:lats_comp}
\end{minipage}
\end{figure}

\begin{table}[t]
\centering
\caption{User study on four different aspects of image aging comparing CUSP.}
\label{tab:userstudy}
\resizebox{0.9\textwidth}{!}{%
\begin{tabular}{lccc|ccc|ccc|c}
    \toprule
      & \multicolumn{3}{c}{\textit{Age accuracy}}        & \multicolumn{3}{c}{\textit{Identity preservation}} & \multicolumn{3}{c}{\textit{Overall quality}}     & \textit{Natural progression} \\
      \midrule
      & \textit{20-29} & \textit{50-69} & \textit{Added} & \textit{20-29}  & \textit{50-69}  & \textit{Added} & \textit{20-29} & \textit{50-69} & \textit{Added} &            -   \\
      \midrule
CUSP  & \textbf{60.2}  & \textbf{72.9}  & \textbf{66.6}  & \textbf{50.8}   & \textbf{63.7}   & \textbf{57.3}  & \textbf{55.8}  & \textbf{67.7}  & \textbf{61.8}  & \textbf{60.6}                \\
HRFAE \cite{hrfae2021} & 17.5           & 15.6           & 16.6           & 24.4            & 24.0            & 24.2           & 21.7           & 20.6           & 21.1           & 24.9                         \\
LATS  \cite{orel2020lifespan} & 22.3           & 11.5           & 16.9           & 24.8            & 12.3            & 18.5           & 22.5           & 11.7           & 17.1           & 14.5                  \\
\bottomrule
\end{tabular}%
}
\end{table}

\noindent\textbf{User study.}
We conducted a study on 80 different users comparing CUPSP with HRFAE \cite{hrfae2021} and LATS \cite{orel2020lifespan}) on the young-to-old and old-to-young tasks on FFHQ-RR. Similarly to \cite{orel2020lifespan}, we asked about user preferences regarding identity preservation, target age accuracy, realism, the naturalness of the age transition, and overall preference.

As seen in Table \ref{tab:userstudy}, CUSP outperforms HRFAE \cite{hrfae2021} and LATS \cite{orel2020lifespan} in every single category by a large margin (CUSP was selected globally in 62\% of cases, compared to 22\% and 17\%, respectively). Furthermore, CUSP's results depict people of the target age with greater accuracy while maintaining the source image identity. On top of that, it outputs higher quality images, and the progression seems more natural and realistic.

\noindent\textbf{Qualitative comparison.}
In Fig.~\ref{fig:hrfae_sota}, we show a qualitative comparison with the state-of-the-art 
evaluated on the \emph{celebA-HQ} dataset, where we transform the input image to the age of 60 years old. First, we observe that Fader, PAGGAN, and IPCGAN generate images with important artifacts. On the contrary, HRFAE, LATS, and our approach generate consistent images with only minor artifacts. However, only CUSP produces images that correspond to the correct target age. Other methods generate images where people look younger than expected since they are unable to make suitable structural changes. Furthermore, LATS operates only in the foreground, requiring a previous masking procedure; for this reason, in Fig.~\ref{fig:hrfae_sota}, the outputs related to LATS display a constant gray background. In addition, CUSP can preserve identity and non-age-related details.

We also perform a qualitative comparison with the two main competitors:  HRFAE on \emph{FFHQ-RR} in Fig. \ref{fig:hrfae_comp} and with LATS on \emph{FFHQ-LS} in Fig. \ref{fig:lats_comp}. We show that CUSP achieves more profound facial structure modifications (\eg thin face shapes that grow wider and wrinkled skin) and hair color transformation. The age progression is smooth. Close ages produce almost identical pictures, but global age progression seems realistic and natural. Regarding LATS (Fig.~\ref{fig:lats_comp}), we see that we obtain similar performance while our method has four major advantages: (1) it operates directly on the entire image and deals with backgrounds and clothing; (2) it does not require an externally trained image segmentation network; (3) CUSP employs a single network while LATS uses a separate network for each gender; and (4) it offers user control as shown in our ablation study (see Sec.~\ref{sec:ablation}). 

\begin{table}[t]
\centering
\caption{Quantitative comparison on \textit{CelebA-HQ} for the \textit{Young} $\rightarrow$ \textit{Old} task employing a target age of 60. CUSP HP (High preservation) is run with $\sigma_m=\sigma_g=1.8$. }
\label{tab:hrfae_comp}

\begin{tabular}{llccccc}
\toprule
\textbf{Method}                         & \textbf{Predicted Age}             & \multicolumn{1}{l}{\textbf{Blur}} & \multicolumn{1}{l}{\textbf{Gender }} & \multicolumn{1}{l}{\textbf{Smiling}} & \multicolumn{1}{l}{\textbf{Neutral}} & \multicolumn{1}{l}{\textbf{Happy}} \\ \midrule
\textit{Real images} & \textit{68.23 $\pm$ 6.54}  &  2.40 &-&-&-&- \\
\midrule
FaderNet                       & 44.34 $\pm$ 11.40            & 9.15                     & 97.60                  & 95.20                           & 90.60                           & 92.40                         \\
PAGGAN                         & 49.07 $\pm$ 11.22            & 3.68                     & 95.10                           & 93.10                           & 90.20                           & 91.70                         \\
IPCGAN                         & 49.72 $\pm$ 10.95            & 9.73                     & 96.70                           & 93.60                           & 89.50                           & 91.10                         \\
HRFAE                          & 54.77 $\pm$ 8.40             & \textbf{2.15}                     & 97.10                           & \textbf{96.30}                  & \textbf{91.30}                  & \textbf{92.70}                \\
HRFAE-224                    & 51.87 $\pm$ 9.59             & 5.49                     & \textbf{97.30}                           & 95.50                           & 88.30                           & 92.50                         \\

LATS \hspace{0.5cm}   & 55.33 $\pm$ 9.33 & 4.77 & 96.55 & 92.70 & 83.77 & 88.64 \\
CUSP HP    & \textbf{67.76 $\pm$ 5.38} & 2.53                     & 93.20                           & 88.70                           & 79.80                           & 84.60                         \\ 
\bottomrule
\end{tabular}%

\end{table}

\begin{table}[t]
\centering
\caption{Quantitative comparison with LATS on the FFHQ-LS dataset for the \emph{age group comparison} task. CUSP CP (Custom preservation) and LP (Low preservation) are run with $(\sigma_m,\sigma_g)=(8,4.5)$ and $(\sigma_m,\sigma_g)=(8,8)$ respectively. }
\label{tab:lats}
\resizebox{0.9\columnwidth}{!}{%

\centering

\begin{tabular}{lccccccc|ccccccc}
    \toprule
                              
                               &\multicolumn{7}{c|}{\textbf{Age MAE}} &\multicolumn{7}{c}{\textbf{Gender Preservation (\%)}}\\
                               \midrule
                                & \textbf{0-2}           & \textbf{3-6}            & \textbf{7-9}            & \textbf{15-19}         & \textbf{30-39}         & \textbf{50-69}         & \textbf{Mean}          &  \textbf{0-2}           & \textbf{3-6}            & \textbf{7-9}            & \textbf{15-19}         & \textbf{30-39}         & \textbf{50-69}         & \textbf{Mean}\\ 
    \midrule
    LATS                       & 7.68 & 8.91 & 6.59 & 5.19 & \textbf{8.23} & \textbf{5.73} & 7.05   & 72.2 & 70.6 & 74.2 & \bf 93.7 & \bf 93.9 & \bf 93.9 & \bf 83.1 \\
    CUSP CP   & 6.89 & \textbf{8.26} & 7.67 & 6.70 & 10.67 & 10.86 & 8.51 & \bf 74.5 & 69.3 & \bf 78.1 & 88.3 & 92.1 & 85.9 & 81.4                 \\
    CUSP LP    & \bf 6.49 & 9.29 & \textbf{5.59} & \textbf{4.99} & 8.36 & 5.74 & \textbf{6.74}   & 69.0 & \bf 76.0 & \bf 78.1 & 87.4 & 86.1 & 80.1 & 79.4\\
    
    \bottomrule
    \end{tabular}%
}
\end{table}
\noindent\textbf{Quantitative comparison.} In Table \ref{tab:hrfae_comp}, we report a quantitative comparison evaluated on the \textit{CelebA-HQ} dataset employing  the \textit{Young} $\rightarrow$ \textit{Old} task. Every model has been trained on \textit{FFHQ-RR}. Regarding HRFAE, we report the performance obtained with models trained and tested at 224$\times$224 and 1024$\times$1024 resolutions (referred to as HRFAE-224 and HRFAE, respectively). We used the available code for LATS to train a model on this dataset. We also report (first row) the mean age predicted by the Face++ classifier when feeding the images of the age class 60 according to the DEX classifier used at training time. We observe an 8.23-year discrepancy. In other words, to generate images that look similar to those labeled as 60 at training time, we need to predict images that the Face++ classifier will perceive on average as 68.23 years old. These experiments confirm that CUSP outperforms other methods, being the only method that substantially modifies the image to adjust the person's target age. 

In addition, CUSP ranks second in terms of Blur, quantifying the good quality of our images. For instance, the performance of HRFAE-224 worsens the predicted age with respect to its 1024$\times$1024 counterpart and deteriorates noticeably in the Blur metric, suggesting a severe drop in the generated image quality.  Interestingly, the more profound and realistic transformations yielded by CUSP and LATS imply slightly worse scores according to the preservation metrics. 
Indeed, preservation metrics suffer from the increased ability to make structural changes to pictures. However, this drop in quantitative fidelity is not manifested in the user study or qualitative results (Figs.~\ref{fig:lats_comp} and \ref{fig:hrfae_comp}). Two hypotheses can explain this discrepancy between qualitative and quantitative results. First, several biases can impact the results (\eg sports clothing is replaced for formal clothes at higher ages, and glasses appear in older targets as well). In addition, there may also be some expression-related biases in different age groups. Second, the CUSP module more frequently targets the image's mouth and eye areas. Those areas are the most related to facial expression detection, and their blurring might negatively affect facial expression preservation. 

We report in Table \ref{tab:lats} a comparison with LATS, both trained and evaluated on the \emph{FFHQ-LS} dataset. The results  support the qualitative analysis performed regarding Figs. \ref{fig:hrfae_comp} and \ref{fig:lats_comp}. Our proposed method is on par with LATS performance concerning the aging task and achieves those results while preserving numerous image details. CUSP with low preservation even outperforms LATS in terms of Mean Age-MAE. We also notice that our approach obtains similar performance in terms of gender preservation while employing a single model and not using gender annotations as in \cite{orel2020lifespan}.  

\subsection{Ablation study}
\label{sec:ablation}

\noindent\textbf{Architecture ablation.} 
We consider four variants of our approach where we ablate the skip connections and the style encoder\footnotemark. In (i), the style encoder is not used; an \emph{Average Pooling layer}replaces $E_s$ on top of the output from $E_c$. 
\textit{(ii)} employs a style encoder but no skip connections,
while \textit{(iii)} employs skip connections in every layer. Finally, \textit{(iv)} follows the proposed architecture employing skip connections in the second-to-last layer only.
In order to make an unbiased evaluation of the architecture and not the masking operation performed by CUSP, we report the performance of CUSP with high preservation $(\sigma_m,\sigma_g) = (7.1,0.0)$, as \textit{(ii)} applies no masking.

\begin{table}[h]
\centering

\parbox{.48\linewidth}{\caption{Ablation study: impact of the skip connections (\emph{SC.}) and the style encoder.}
\label{tab:ablation_arch}

{%
\resizebox{0.48\textwidth}{!}{\begin{tabular}{lccc}
\toprule

                                & \multicolumn{1}{l}{\bf LPIPS}             & \bf Age MAE         & \bf Mean KID        \\ \midrule
\textit{(i)} No style encoder       & 0.84                                         & 6.21                      & 0.0163          \\
\textit{(ii)} No \emph{SC.}       & 1.70                                         & \bf 6.17                      & 0.0109  \\        
\textit{(iii)} \emph{SC.} at every layer & 1.85 & 6.34 & 0.0175 \\

\textit{(iv)} Full & \bf 0.78                                         & 6.29                      & \bf 0.0089          \\ \bottomrule
\end{tabular}}}%
}

\parbox{.48\linewidth}{\caption{Ablation study: impact of the masking strategy used in CUSP.}
\label{tab:ablation_mask}

{%
\resizebox{0.48\textwidth}{!}{\begin{tabular}{lccc}
\toprule

                                & \multicolumn{1}{l}{\bf LPIPS}             & \bf Age MAE         & \bf Mean KID        \\ \midrule
         
 Top-class GB & 1.25 & 6.19 &  0.0145 \\
\
Class-indep. (Ours)& \bf 0.78                                         & \bf 6.29                      & \bf 0.0089          \\ \bottomrule
\end{tabular}}}%

}
\end{table}

Results shown in Table~\ref{tab:ablation_arch} suggest that a separate style encoder, as in our \emph{Full} model \emph{(iv)}, yields better reconstruction (lower LPIPS) and similar aging performance (Age MAE and Mean KID) than using a single encoder for both content and style as in \emph{(i)}.
Regarding skip connections, not using them leads to an important reconstruction error (see high LPIPS) since the network cannot reconstruct the image details.
 However, skip connections in every layer also results in low reconstruction performance. We hypothesize that the model faces optimization issues. More specifically, adding skip connections on every layer dramatically increases the decoder's complexity (approximately doubling its number of parameters), making the network slower and harder to
train.

\begin{figure}[htb]
\begin{minipage}{0.68\linewidth}
\centering
\includegraphics[width=\linewidth]{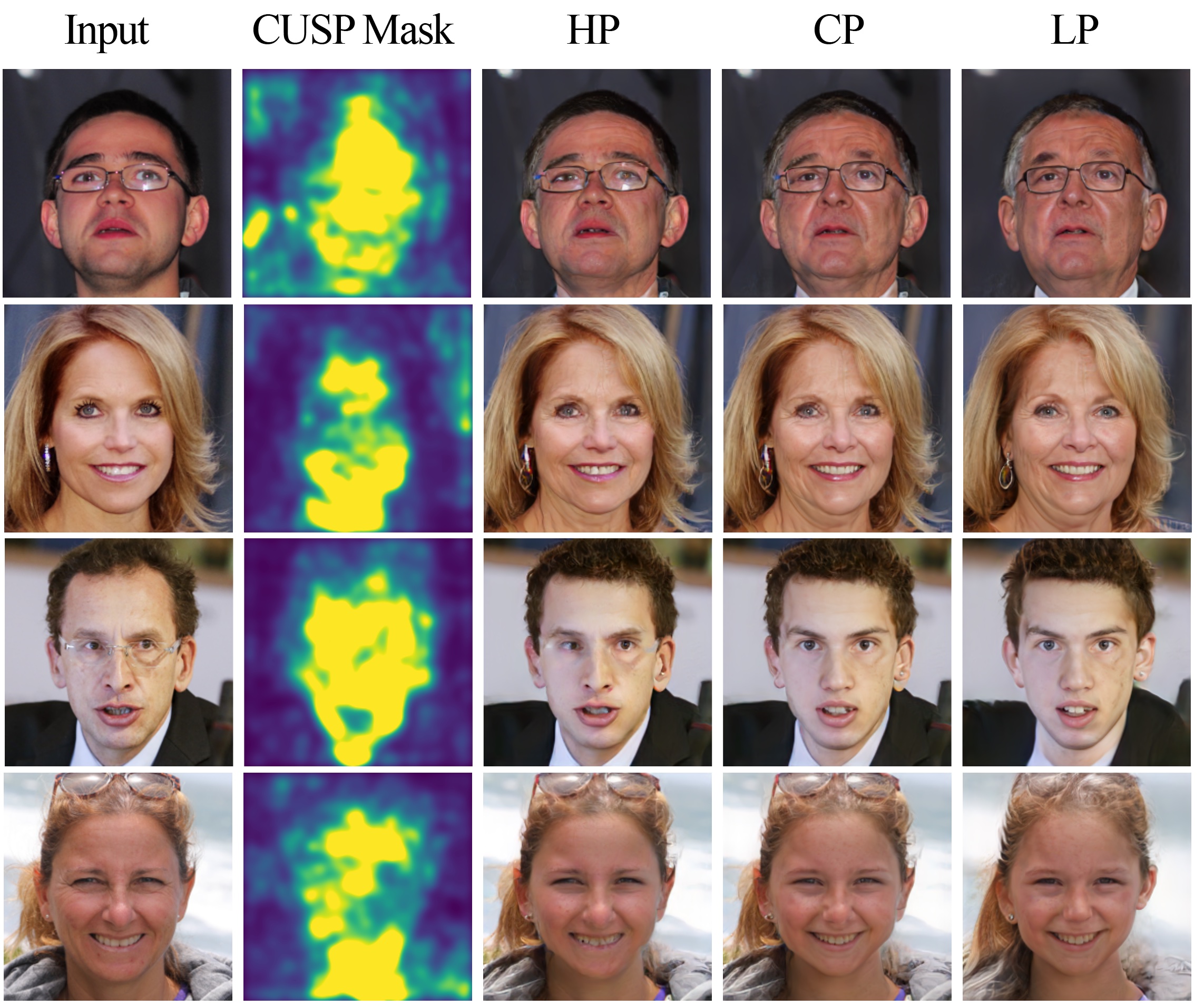}
  \caption{Impact of the kernel value: images obtained with High, Low, and Custom structure preservation (LP, HP, and CP).   HP:$(\sigma_m,\sigma_g)=(0,0)$; CP:$(\sigma_m,\sigma_g)=(9,0)$; HP:$(\sigma_m,\sigma_g)=(9,9)$. The second column shows the mask estimated by CUSP.}
  \label{fig:qualitative_ablation_cups}
  
\end{minipage}
\hfill
\begin{minipage}{0.28\linewidth}

\centering

\includegraphics[width=\linewidth]{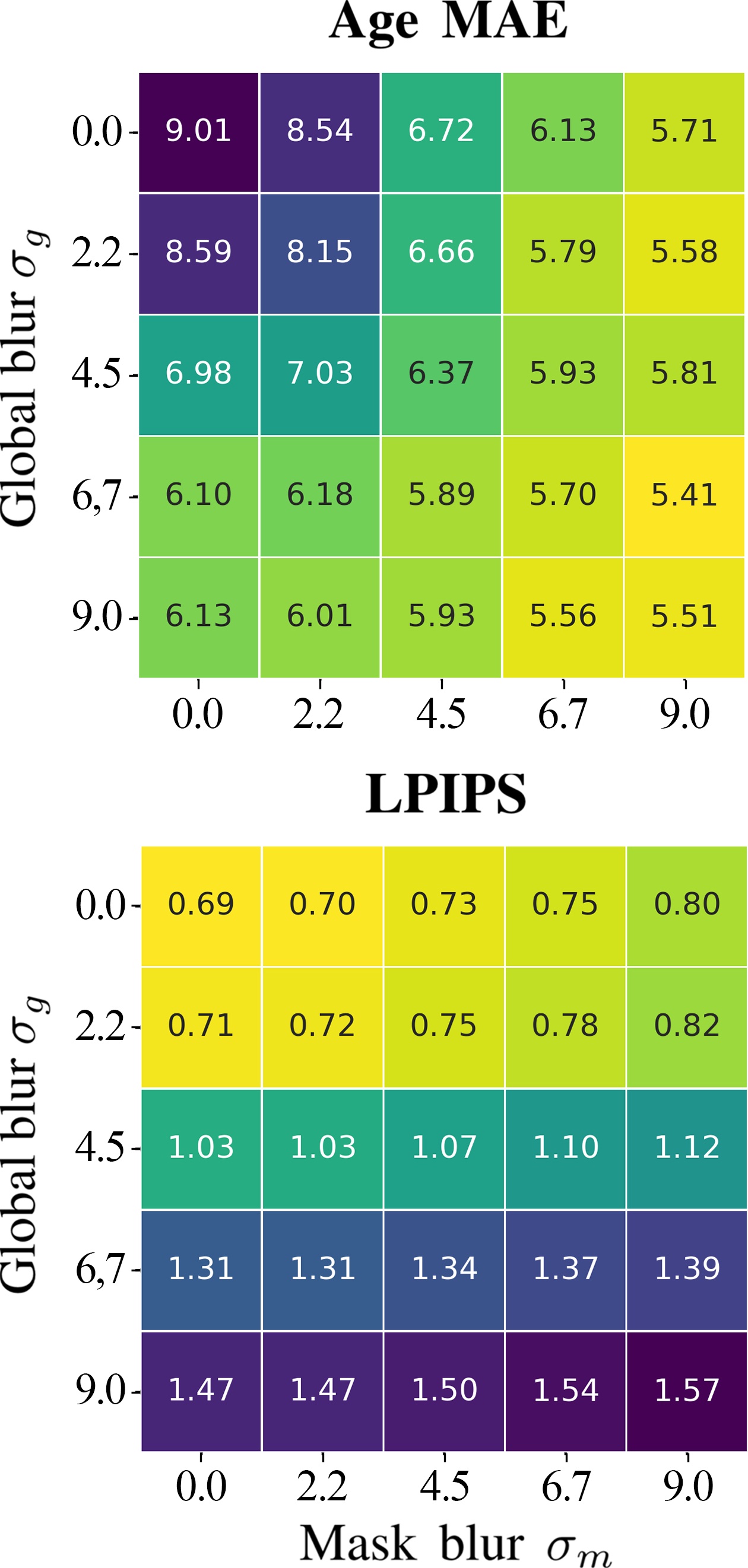}
\caption{CUSP parameters and impact on Age MAE (left) and LPIPS$\times 10$ (right).}
\label{fig:ablation_blur}

\end{minipage}

\end{figure}

\noindent\textbf{CUSP module analysis.} In Fig. \ref{fig:qualitative_ablation_cups}, we qualitatively evaluate the impact of the kernel values used in CUSP. 
We compare images obtained with Low, Custom, and High structure preservation (referred to as LP, CP, and HP), where we use kernel values ranging from $\sigma=0$ to $\sigma=9$. We also display the mask $\Mmat$ estimated by the CUSP module. We observe that when the user provides low kernel values (\ie higher preservation), the shape of the face is kept, while with higher kernel values, the network has the freedom to change its shape. The impact is clearly visible on the neck and chin of the women in the second and last row.

The visualization of the mask shows that our approach identifies those regions that change with age (chin, mouth, and forehead).  We also quantitatively measure the impact of each kernel parameter.
In Fig. \ref{fig:ablation_blur}, we report the Age MAE and LPIPS while changing the local and global blur parameters. By increasing the local blur, we can see that CUSP achieves a significantly lower age error while keeping a small reconstruction error. On the contrary, using global blur to improve the age performance (\ie reduce the age MAE) implies a substantial increase in the LPIPS metric, reflecting some loss of details.  Overall, these experiments demonstrate the conflicting nature of aging and reconstruction performances. 
These observations further justify our motivation to offer the user the possibility of controlling this trade-off, thereby demonstrating the value of CUSP and its masking strategy. The ability to modify both $\sigma_m$ and $\sigma_g$ with different values allows us to achieve the same age-accurate transformation results while minimizing the reconstruction performance drop.

We complete this analysis with an ablation study regarding the GB-based computation of the CUSP masks. More precisely, two strategies are compared: in \emph{Top-1 class}, we apply GB on the most-activated class, while in \emph{class-independent}, we adopt the proposed strategy of taking the sum of the classification layer before softmax. Results reported in Tab. \ref{tab:ablation_mask} demonstrate that the class-independent strategy performs best. Indeed, using every class output from the age classifier might benefit the masking, as every age-related feature is relevant for the translation, not only those involving its current age.


\section{Conclusions}

We present a novel architecture for face age editing that can produce structural facial modifications while preserving relevant details in the original image. Our proposal has two main contributions. First, we propose a style-based strategy to combine the style and content representations of the input image while conditioning the output on the target age. Second, we present a Custom Structure Preservation (CUSP) module that allows users to adjust the degree of structure preservation in the input image at inference time. We validate our approach by comparing six state-of-the-art solutions and employing three datasets. 
Our results suggest that our method generates more natural-looking, age-accurate transformed images and allows more profound facial changes while adequately preserving identity and modifying only age-related aspects. An extensive user study further confirmed this analysis.
We plan to extend CUSP to other image editing tasks in future works.

\par\vfill\par

\clearpage

\bibliographystyle{splncs04}
\bibliography{egbib}

\begin{thebibliography}{10}
\providecommand{\url}[1]{\texttt{#1}}
\providecommand{\urlprefix}{URL }
\providecommand{\doi}[1]{https://doi.org/#1}

\bibitem{ak2019attribute}
Ak, K.E., Lim, J.H., Tham, J.Y., Kassim, A.A.: Attribute manipulation
  generative adversarial networks for fashion images. In: IEEE/CVF ICCV (2019)

\bibitem{alaluf2021matter}
Alaluf, Y., Patashnik, O., Cohen-Or, D.: Only a matter of style: Age
  transformation using a style-based regression model. ACM Trans. Graph.
  \textbf{40}(4) (2021)

\bibitem{antipov2017face}
Antipov, G., Baccouche, M., Dugelay, J.L.: Face aging with conditional
  generative adversarial networks. In: IEEE ICIP (2017)

\bibitem{binkowski2018kid}
Bi{\'n}kowski, M., Sutherland, D.J., Arbel, M., Gretton, A.: Demystifying mmd
  gans. arXiv preprint arXiv:1801.01401  (2018)

\bibitem{chen2017rethinking}
Chen, L.C., Papandreou, G., Schroff, F., Adam, H.: Rethinking atrous
  convolution for semantic image segmentation. arXiv preprint arXiv:1706.05587
  (2017)

\bibitem{choi2018stargan}
Choi, Y., Choi, M., Kim, M., Ha, J.W., Kim, S., Choo, J.: Stargan: Unified
  generative adversarial networks for multi-domain image-to-image translation.
  In: IEEE/CVF CVPR (2018)

\bibitem{fu2019geometry}
Fu, H., Gong, M., Wang, C., Batmanghelich, K., Zhang, K., Tao, D.:
  Geometry-consistent generative adversarial networks for one-sided
  unsupervised domain mapping. In: IEEE/CVF CVPR (2019)

\bibitem{fu2010age}
Fu, Y., Guo, G., Huang, T.S.: Age synthesis and estimation via faces: A survey.
  IEEE T-PAMI  (2010)

\bibitem{He2019S2GANSA}
He, Z., Kan, M., Shan, S., Chen, X.: S2gan: Share aging factors across ages and
  share aging trends among individuals. IEEE/CVF ICCV  (2019)

\bibitem{heusel2017fid}
Heusel, M., Ramsauer, H., Unterthiner, T., Nessler, B., Hochreiter, S.: Gans
  trained by a two time-scale update rule converge to a local nash equilibrium.
  Neurips  \textbf{30} (2017)

\bibitem{huang2018multimodal}
Huang, X., Liu, M.Y., Belongie, S., Kautz, J.: Multimodal unsupervised
  image-to-image translation. In: IEEE/CVF ECCV (2018)

\bibitem{isola2017image}
Isola, P., Zhu, J.Y., Zhou, T., Efros, A.A.: Image-to-image translation with
  conditional adversarial networks. In: IEEE/CVF CVPR (2017)

\bibitem{Karras2019stylegan2}
Karras, T., Laine, S., Aittala, M., Hellsten, J., Lehtinen, J., Aila, T.:
  Analyzing and improving the image quality of stylegan. In: IEEE Conf. Comput.
  Vis. Pattern Recog. (2020)

\bibitem{karras2017pggan}
Karras, T., Aila, T., Laine, S., Lehtinen, J.: Progressive growing of gans for
  improved quality, stability, and variation. ICLR  (2017)

\bibitem{karras2020training}
Karras, T., Aittala, M., Hellsten, J., Laine, S., Lehtinen, J., Aila, T.:
  Training generative adversarial networks with limited data. arXiv preprint
  arXiv:2006.06676  (2020)

\bibitem{stylegan1}
Karras, T., Laine, S., Aila, T.: A style-based generator architecture for
  generative adversarial networks. In: IEEE/CVF CVPR (2019)

\bibitem{stylegan2}
Karras, T., Laine, S., Aittala, M., Hellsten, J., Lehtinen, J., Aila, T.:
  Analyzing and improving the image quality of stylegan. In: IEEE/CVF CVPR
  (2020)

\bibitem{kemelmacher2014illumination}
Kemelmacher-Shlizerman, I., Suwajanakorn, S., Seitz, S.M.: Illumination-aware
  age progression. In: IEEE/CVF CVPR (2014)

\bibitem{Kim_2021_CVPR}
Kim, D., Khan, M.A., Choo, J.: Not just compete, but collaborate: Local
  image-to-image translation via cooperative mask prediction. In: IEEE/CVF CVPR
  (2021)

\bibitem{lample2017fader}
Lample, G., Zeghidour, N., Usunier, N., Bordes, A., DENOYER, L., et~al.: Fader
  networks: Manipulating images by sliding attributes. In: Neurips (2017)

\bibitem{lee2018diverse}
Lee, H.Y., Tseng, H.Y., Huang, J.B., Singh, M., Yang, M.H.: Diverse
  image-to-image translation via disentangled representations. In: IEEE/CVF
  ECCV (2018)

\bibitem{Liu_nips2017}
Liu, M.Y., Breuel, T., Kautz, J.: Unsupervised image-to-image translation
  networks. In: Neurips (2017)

\bibitem{liu2015deep}
Liu, Z., Luo, P., Wang, X., Tang, X.: Deep learning face attributes in the
  wild. In: IEEE/CVF ICCV (2015)

\bibitem{reaging2021}
Makhmudkhujaev, F., Hong, S., Park, I.K.: Re-aging gan: Toward personalized
  face age transformation. In: IEEE/CVF ICCV (2021)

\bibitem{miyato2018cgans}
Miyato, T., Koyama, M.: cgans with projection discriminator. arXiv preprint
  arXiv:1802.05637  (2018)

\bibitem{muhammad2020eigen}
Muhammad, M.B., Yeasin, M.: Eigen-cam: Class activation map using principal
  components. In: 2020 International Joint Conference on Neural Networks
  (IJCNN). pp.~1--7. IEEE (2020)

\bibitem{orel2020lifespan}
Or-El, R., Sengupta, S., Fried, O., Shechtman, E., Kemelmacher-Shlizerman, I.:
  Lifespan age transformation synthesis. In: IEEE/CVF ECCV (2020)

\bibitem{pan2018mean}
Pan, H., Han, H., Shan, S., Chen, X.: Mean-variance loss for deep age
  estimation from a face. In: IEEE/CVF CVPR (2018)

\bibitem{park2020swapping}
Park, T., Zhu, J.Y., Wang, O., Lu, J., Shechtman, E., Efros, A., Zhang, R.:
  Swapping autoencoder for deep image manipulation. Advances in Neural
  Information Processing Systems  \textbf{33},  7198--7211 (2020)

\bibitem{pumarola2018ganimation}
Pumarola, A., Agudo, A., Martinez, A.M., Sanfeliu, A., Moreno-Noguer, F.:
  Ganimation: Anatomically-aware facial animation from a single image. In:
  IEEE/CVF ECCV (2018)

\bibitem{richardson21cvpr}
Richardson, E., Alaluf, Y., Patashnik, O., Nitzan, Y., Azar, Y., Shapiro, S.,
  Cohen-Or, D.: Encoding in style: a {StyleGAN} encoder for image-to-image
  translation. In: IEEE Conf. Comput. Vis. Pattern Recog. (2021)

\bibitem{unet2015}
Ronneberger, O., Fischer, P., Brox, T.: U-net: Convolutional networks for
  biomedical image segmentation. In: MICCAI. Springer (2015)

\bibitem{rothe2018dex}
Rothe, R., Timofte, R., Gool, L.V.: Deep expectation of real and apparent age
  from a single image without facial landmarks. International Journal of
  Computer Vision  \textbf{126}(2-4),  144--157 (2018)

\bibitem{dex2015}
Rothe, R., Timofte, R., Van~Gool, L.: Dex: Deep expectation of apparent age
  from a single image. In: IEEE/CVF ICCV-W (2015)

\bibitem{gradcam2017}
Selvaraju, R.R., Cogswell, M., Das, A., Vedantam, R., Parikh, D., Batra, D.:
  Grad-cam: Visual explanations from deep networks via gradient-based
  localization. In: IEEE/CVF ICCV (2017)

\bibitem{siarohin2018deformable}
Siarohin, A., Sangineto, E., Lathuiliere, S., Sebe, N.: Deformable gans for
  pose-based human image generation. In: IEEE/CVF CVPR (2018)

\bibitem{guidedbackprop2015}
Springenberg, J.T., Dosovitskiy, A., Brox, T., Riedmiller, M.A.: Striving for
  simplicity: The all convolutional net. In: 3rd International Conference on
  Learning Representations, {ICLR} 2015, San Diego, CA, USA, May 7-9, 2015,
  Workshop Track Proceedings (2015)

\bibitem{srinivas2019full}
Srinivas, S., Fleuret, F.: Full-gradient representation for neural network
  visualization. Advances in neural information processing systems  \textbf{32}
  (2019)

\bibitem{tang2019attention}
Tang, H., Xu, D., Sebe, N., Yan, Y.: Attention-guided generative adversarial
  networks for unsupervised image-to-image translation. In: IJCNN (2019)

\bibitem{wang2016recurrent}
Wang, W., Cui, Z., Yan, Y., Feng, J., Yan, S., Shu, X., Sebe, N.: Recurrent
  face aging. In: IEEE/CVF CVPR (2016)

\bibitem{wang2018face}
Wang, Z., Tang, X., Luo, W., Gao, S.: Face aging with identity-preserved
  conditional generative adversarial networks. In: IEEE/CVF CVPR (2018)

\bibitem{yang2018learning}
Yang, H., Huang, D., Wang, Y., Jain, A.K.: Learning face age progression: A
  pyramid architecture of gans. In: IEEE/CVF CVPR (2018)

\bibitem{Yao_2021_ICCV}
Yao, X., Newson, A., Gousseau, Y., Hellier, P.: A latent transformer for
  disentangled face editing in images and videos. In: IEEE/CVF ICCV (2021)

\bibitem{hrfae2021}
Yao, X., Puy, G., Newson, A., Gousseau, Y., Hellier, P.: High resolution face
  age editing. In: IEEE ICPR (2021)

\bibitem{lpips2018}
Zhang, R., Isola, P., Efros, A.A., Shechtman, E., Wang, O.: The unreasonable
  effectiveness of deep features as a perceptual metric. In: CVPR (2018)

\bibitem{zhang2017age}
Zhang, Z., Song, Y., Qi, H.: Age progression/regression by conditional
  adversarial autoencoder. In: IEEE/CVF CVPR (2017)

\bibitem{zhu2017unpaired}
Zhu, J.Y., Park, T., Isola, P., Efros, A.A.: Unpaired image-to-image
  translation using cycle-consistent adversarial networks. In: IEEE/CVF ICCV
  (2017)

\bibitem{zhu2017multimodal}
Zhu, J.Y., Zhang, R., Pathak, D., Darrell, T., Efros, A.A., Wang, O.,
  Shechtman, E.: Multimodal image-to-image translation by enforcing bi-cycle
  consistency. In: Neurips (2017)

\end{thebibliography}

\title{Supplementary Material:\\ Custom Structure Preservation in Face Aging} 

\titlerunning{Custom Structure Preservation in Face Aging}

\author{Guillermo Gomez-Trenado\inst{1}\orcidlink{0000-0003-3366-6047} \and
Stéphane Lathuilière\inst{2}\orcidlink{0000-0001-6927-8930} \and
Pablo Mesejo\inst{1} \and
Óscar Cordón\inst{1}}
\authorrunning{Gomez-Trenado et al.}

\institute{
DaSCI research institute, DECSAI, University of Granada, Granada, Spain \email{\{guillermogomez,pmesejo,ocordon\}@ugr.es}
\and
LTCI, Télécom-Paris, Intitute Polytechnique de Paris, Palaiseau, France
\email{stephane.lathuiliere@telecom-paris.fr}}

\maketitle

In these supplementary materials, we provide additional experiments and visual examples that show the performance of our proposal. First, in Sec.~\ref{sec:AMAE}, we provide experiments to illustrate the DEX-Face++ age misalignment mentioned in the main paper. Second, in Sec.~\ref{sec:arch}, we further describe our proposed architecture. Third, in Sec.~\ref{sec:userstudy}, we describe the user study in detail. In Sec.~\ref{sec:ablation}, we provide additional qualitative examples, including failure cases and a discussion of examples to complete the ablation study of the main paper. Then, in Sec.~\ref{sec:sota}, we complete our comparison with the State of the Art, including additional results. Finally, in Sec. \ref{sec:licence}, we list the licenses of the datasets used in our experiments.

\section{Age estimation correction}

\label{sec:AMAE}
Our evaluation protocol uses Face++ to estimate the person's age in the generated image. However, as mentioned in the main paper, the misalignment of DEX and Face++ classifiers may bias evaluation. In this section, we illustrate how the DEX-Face++ misalignment can bias evaluation.

 \begin{figure}[h!]\centering\centering
\includegraphics[width=0.4\textwidth]{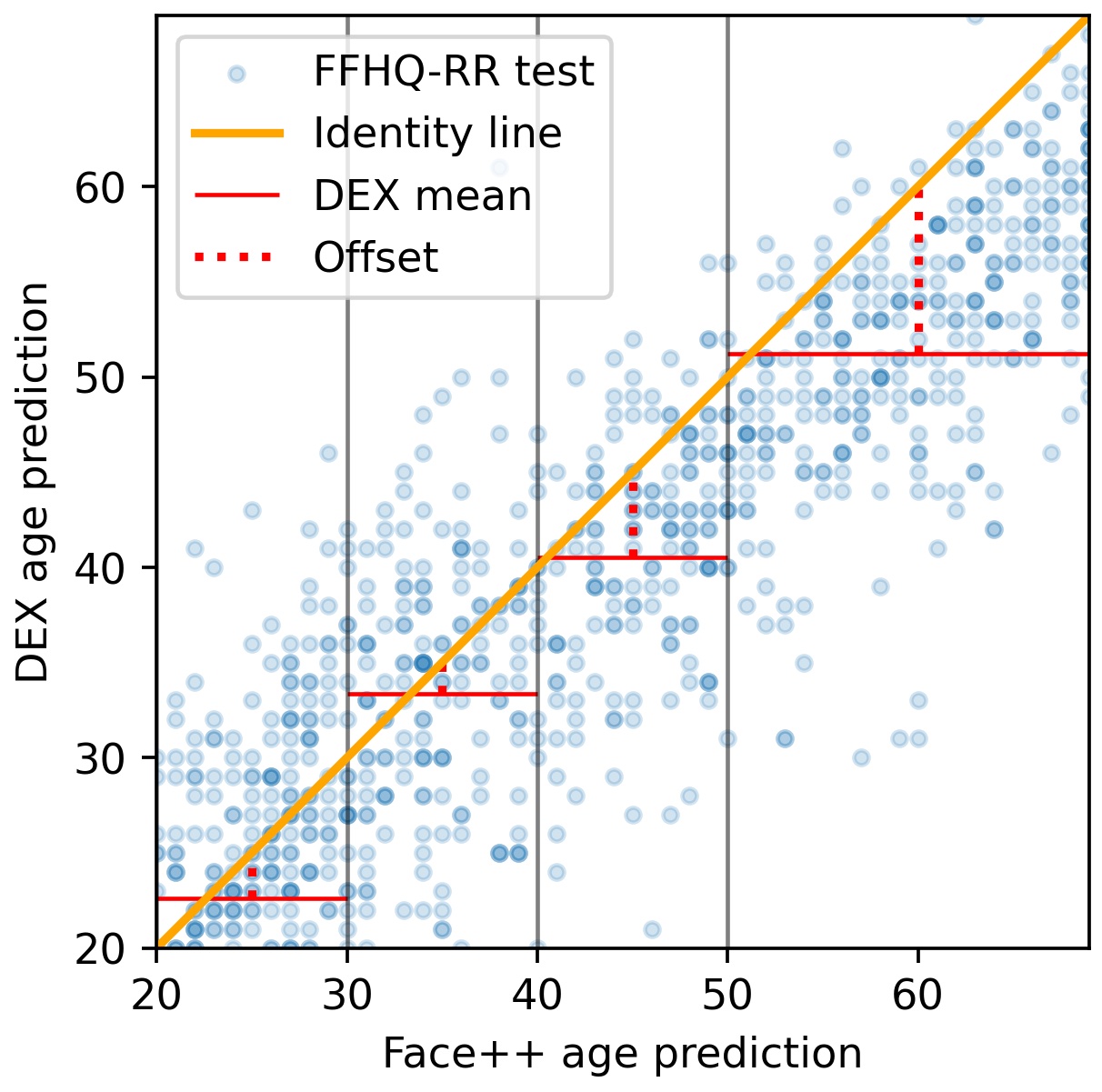}
  \caption{DEX classifier and Face++ distribution discrepancy by age group on \textit{FFHQ-RR} test set. Color intensity denotes distribution density. The red horizontal lines represent the mean age of each age group according to DEX.}
  \label{fig:dex_facepp}
\end{figure}
 As in previous approaches~\cite{hrfae2021,zhang2017age}, DEX is used at training time on the \emph{FFHQ-RR} dataset. Thus, the aging task consists in generating images that match DEX predictions.
 The DEX-Face++ discrepancy may bias evaluation since an aging method that fails in generating images corresponding to the target age could be favored if the method is biased in the same direction as the Face++ classifier. 
 
 To visualize this discrepancy, we plot in Fig~\ref{fig:dex_facepp} the distribution of the  DEX-Face++ predictions on the FFHQ-RR dataset. In the case of perfect agreement, all the blue points would be located on the orange identity line. We also report the mean age of each age group according to DEX (red horizontal lines). A vertical dotted line represents the amplitude of the discrepancy. In this case, the discrepancy is especially noticeable in older groups.
 
 Therefore, in our evaluation protocol, we estimate the age of the original images with Face++ and compute the mean for each group. Age MAE is then computed as the distance between the mean group predicted age and the transformed image predicted age.

\section{Age modulation architecture}
\label{sec:arch}

\begin{figure}[h]
    \centering
    \includegraphics[width=0.6\textwidth]{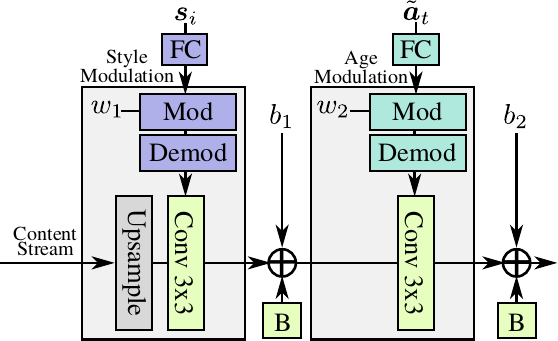}
    \caption{Illustration of the decoder blocks used in $G$ . ~\framebox{B} denotes the noise broadcast operation, and \framebox{FC} denotes a fully-connected layer. $w_1$ and $w_2$ are two learned scaling parameters, while $b_1$ and $b_2$ are learned biases.}
    \label{fig:block}
\end{figure}

In the main paper, we describe our decoder architecture.
Its architecture is based on StyleGAN-2 \cite{stylegan2}, which achieves state-of-the-art performance in unconditional image generation. In addition, we provide several modifications to adapt it to the aging task. These are the use of skip connections, the active manipulation of the skip connection through our CUSP module, and the use of two different inputs in each generator block. The former two are thoroughly discussed in our main work. We now further discuss the latter.

An illustration of our decoder block can be found in Figure \ref{fig:block}. Unlike \cite{stylegan2}, our decoder block takes three inputs: the former block output, the style embedding, and the age embedding. Each decoder block outputs an image twice the size of its input and is composed of two consecutive sub-blocks: the \textit{style sub-block} and the \textit{age block}. In the \textit{style sub-block}, the input is upscaled through bilinear interpolation. Then the upscaled input is transformed through weight demodulation ($w_1$) based on a linear combination of the style embedding ($s_i$). In the second sub-block, the age embedding $\tilde{a}_t$ and $w_2$ are used for transforming the former sub-block output. Both $s_i$ and $\tilde{a}_t$ are shared by every block. After each step, 0-centered random noise $B$ is added to the output.

\section{User study}
\label{sec:userstudy}
We now provide some details regarding the user study reported in the main.
Each test consisted of 48 random questions on four different topics. In total, 72 users were evaluated. Similarly to \cite{orel2020lifespan} approach this is the description prompted to the users.

\mybox{
\small
In this study, you will be presented with several sets of images to choose from. We will compare several AI solutions to transform a person's age in an image, similar to widely known apps like FaceApp. There are four kinds of questions, you'll have to click on your chosen image, there are no correct answers:
\begin{enumerate}
    \item \label{item:ageacc} \textbf{Age accuracy:} From the images displayed, which one better depicts a person from the \textit{target age group}? An actual person's picture (not shown) has been transformed to a target age with different mechanisms. We want to know which one you think is more accurate.
    \item \label{item:identity} \textbf{Identity preservation:} From the images displayed, which one better transforms the shown original picture to the target age group while \textit{reasonably maintaining the person's identity}? You'll have to judge which result seems more reasonable, attending to age transformation and identity preservation.
    \item \label{item:overall} \textbf{Overall better:} From the images displayed, which one is \textit{overall better} transforming the age of the person depicted in the picture? Which one do you prefer? Which image seems more pleasing?
    \item \label{item:progression} \textbf{Whole age progression:} From the different shown \textit{age progressions}, which seems \textit{more natural and reasonable}?
\end{enumerate}
In case of doubt, choose the image you subjectively prefer.
}

From FFHQ-RR, 50 images were selected for each group (20-29, 30-39,40-49,50-69) and transformed to target ages 25, 35, 45, 60 with each comparing method (HRFAE \cite{hrfae2021}, LATS \cite{orel2020lifespan}, and ours), resulting in 200 original images and 2400 transformed images. Every image from our method was obtained with CP configuration $(\sigma_m,\sigma_g)=(8,1.8)$. Age translations were done from 20-29 and 30-39 to 50-69 (young to old) and from 40-49 and 50-69 to 20-29 (old to young).

In Question-Kind \ref{item:ageacc} (QK  \ref{item:ageacc}) and QK  \ref{item:overall}, three randomly ordered transformed images were presented next to a target age group. In QK \ref{item:identity}, the original image is included. Finally, in QK \ref{item:progression}, besides the original image, four images showing age progression (25, 35, 45, and 60) are presented for each method.

\section{Ablation Study}
\label{sec:ablation}

\subsection{CUSP processing}

\begin{figure}[t]\centering
\includegraphics[width=0.6\textwidth]{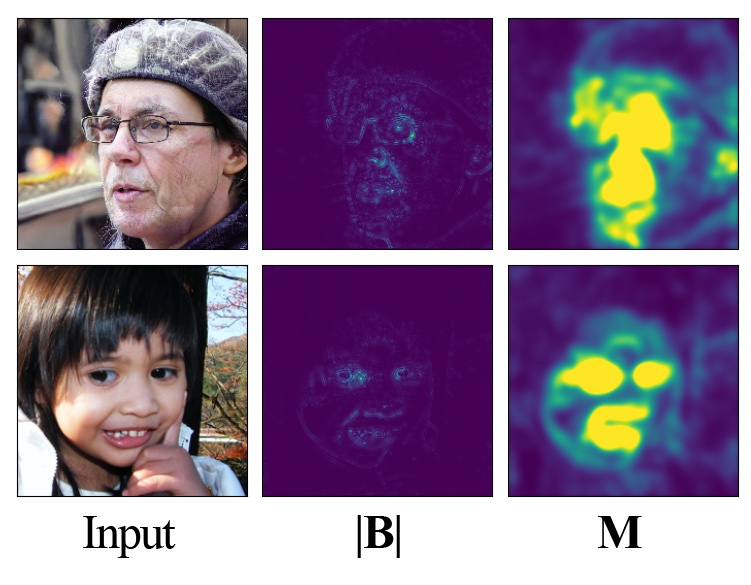}
  \caption{Example outputs of the CUSP module.  From left to right: 1) Input image, 2) Matrix $|\Bmat|$, the absolute value of the guided backpropagation output averaged over the RGB dimension, 3) Mask $\Mmat$ predicted by the CUSP module.}
  \label{fig:cuspout}
\end{figure}

We now provide some visualizations that motivate the proposed computation for the mask $\Mmat$. Figure \ref{fig:cuspout} shows the output $\Bmat$ of the Guided backpropagation algorithm for two input images (2nd column). We see that $\Bmat$ is very sparse. Therefore, we apply blur before normalization and clipping to enlarge the activated regions. In this way, we obtain the mask in the last column. We see that the high values of the masks are primarily located in the eye and mouth regions, while the background is associated with very low values. This visualization shows that our CUSP module can act only on the relevant regions in the foreground. 

\begin{figure}[h]
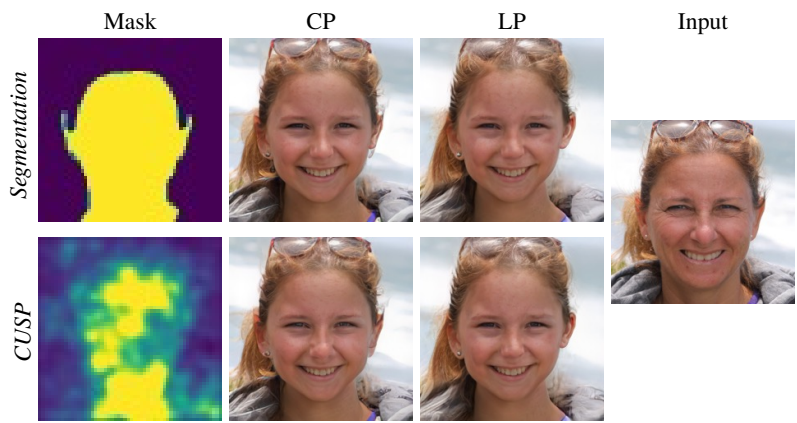

  \centering
  \begin{tabular}{rcccc}
      & Mask & CP & LP & Input \\
  \rotatebox[origin=c]{90}{\textit{Segmentation}} & \adjimage{rebuttal_imgs/custom_mask/\imgid_segm_\imgage_mask.jpg} & \adjimage{rebuttal_imgs/custom_mask/\imgid_segm_\imgage_CP.jpg} & \adjimage{rebuttal_imgs/custom_mask/\imgid_segm_\imgage_LP.jpg} & \multirow{2}{*}{\adjimage{rebuttal_imgs/custom_mask/\imgid.jpg}}\\
  \rotatebox[origin=c]{90}{\textit{CUSP}} &  \adjimage{rebuttal_imgs/custom_mask/\imgid_cusp_\imgage_mask.jpg} & \adjimage{rebuttal_imgs/custom_mask/\imgid_cusp_\imgage_CP.jpg} & \adjimage{rebuttal_imgs/custom_mask/\imgid_cusp_\imgage_LP.jpg} \\
  \end{tabular}
  \caption{Qualitative evaluation of segmentation-based masking. }
  \label{fig:mask_comp}
  \end{figure}

\new{
Furthermore, we compared our CUSP module with supervised alternatives such as segmentation-based masking \cite{chen2017rethinking,orel2020lifespan}. Even though our predicted mask is not always accurate, it has several advantages: \textit{(i)} It rules out the need for extra supervision (\textit{e.g.}, landmarks); \textit{(ii)} CUSP with the Custom Preservation (CP) setting targets only age-specific regions while face segmentation uniformly blurs the whole face area; \textit{(iii)} When we evaluated a segmentation model (\textit{BiSeNet} trained on \textit{CelebAMask-HQ}) as the CUSP mask, it showed that the segmentation-based mask introduces new artifacts (see left ear on Fig. \ref{fig:mask_comp}) probably due to mask inaccuracies. Regarding CUSP,  Low Preservation (LP) setting losses background details, but CP manages to preserve the background despite the inaccurate mask.
}

\subsection{Architecture ablation}

\begin{figure}[t]
    \centering
    \begin{tabular}{llllll}
    \multicolumn{1}{c}{} & \multicolumn{1}{c}{Original} & \multicolumn{1}{c}{25} & \multicolumn{1}{c}{35} & \multicolumn{1}{c}{45} & \multicolumn{1}{c}{60} \\ 
    \makecell{No style \\ encoder} & \includegraphics[width=.16\linewidth,valign=m]{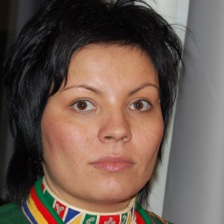} & \includegraphics[width=.16\linewidth,valign=m]{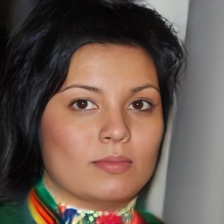} & \includegraphics[width=.16\linewidth,valign=m]{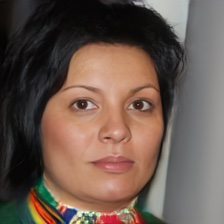} & \includegraphics[width=.16\linewidth,valign=m]{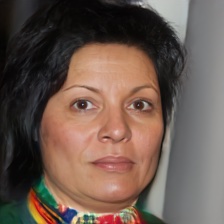} & \includegraphics[width=.16\linewidth,valign=m]{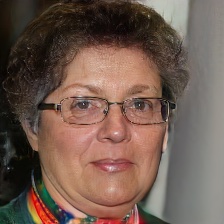}\vspace{0.2cm} \\
\makecell{No SC.} & \includegraphics[width=.16\linewidth,valign=m]{ablation/00718.jpg} & \includegraphics[width=.16\linewidth,valign=m]{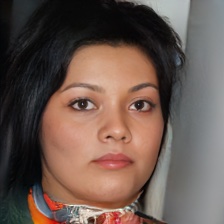} & \includegraphics[width=.16\linewidth,valign=m]{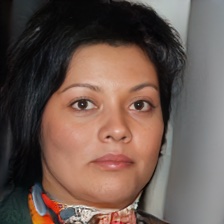} & \includegraphics[width=.16\linewidth,valign=m]{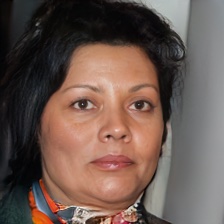} & \includegraphics[width=.16\linewidth,valign=m]{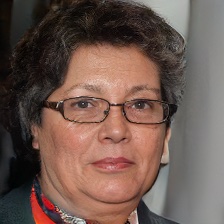}\vspace{0.2cm} \\
\makecell{SC. at \\ every layer} & \includegraphics[width=.16\linewidth,valign=m]{ablation/00718.jpg} & \includegraphics[width=.16\linewidth,valign=m]{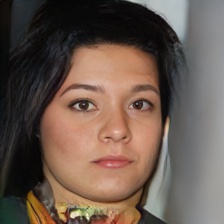} & \includegraphics[width=.16\linewidth,valign=m]{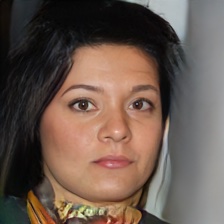} & \includegraphics[width=.16\linewidth,valign=m]{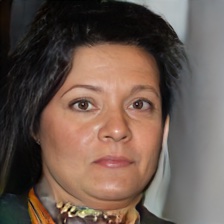} & \includegraphics[width=.16\linewidth,valign=m]{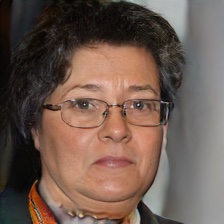}\vspace{0.2cm} \\
\makecell{Full} & \includegraphics[width=.16\linewidth,valign=m]{ablation/00718.jpg} & \includegraphics[width=.16\linewidth,valign=m]{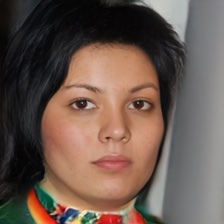} & \includegraphics[width=.16\linewidth,valign=m]{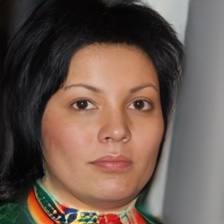} & \includegraphics[width=.16\linewidth,valign=m]{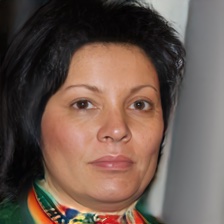} & \includegraphics[width=.16\linewidth,valign=m]{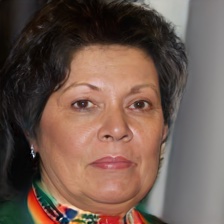}\vspace{0.2cm} \\
    
    \end{tabular}
    
    \begin{tabular}{llllll}
    \multicolumn{1}{c}{} & \multicolumn{1}{c}{Original} & \multicolumn{1}{c}{25} & \multicolumn{1}{c}{35} & \multicolumn{1}{c}{45} & \multicolumn{1}{c}{60} \\ 
    
   \makecell{No style \\ encoder} & \includegraphics[width=.16\linewidth,valign=m]{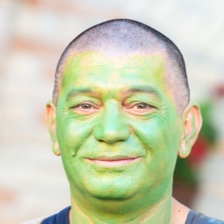} & \includegraphics[width=.16\linewidth,valign=m]{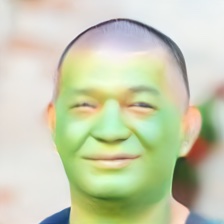} & \includegraphics[width=.16\linewidth,valign=m]{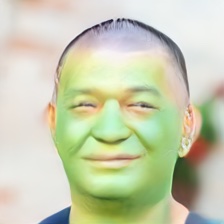} & \includegraphics[width=.16\linewidth,valign=m]{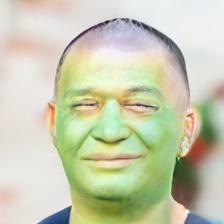} & \includegraphics[width=.16\linewidth,valign=m]{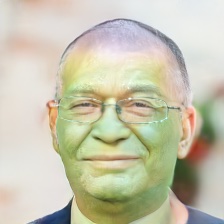}\vspace{0.2cm} \\
\makecell{No SC.} & \includegraphics[width=.16\linewidth,valign=m]{ablation/02492.jpg} & \includegraphics[width=.16\linewidth,valign=m]{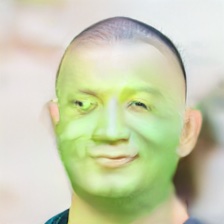} & \includegraphics[width=.16\linewidth,valign=m]{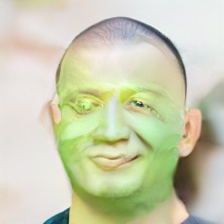} & \includegraphics[width=.16\linewidth,valign=m]{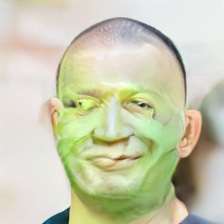} & \includegraphics[width=.16\linewidth,valign=m]{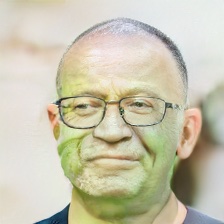}\vspace{0.2cm} \\
\makecell{SC. at \\ every layer} & \includegraphics[width=.16\linewidth,valign=m]{ablation/02492.jpg} & \includegraphics[width=.16\linewidth,valign=m]{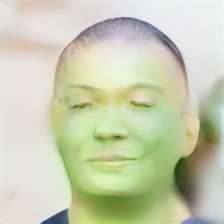} & \includegraphics[width=.16\linewidth,valign=m]{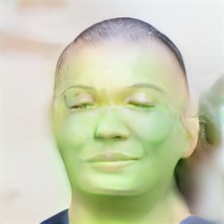} & \includegraphics[width=.16\linewidth,valign=m]{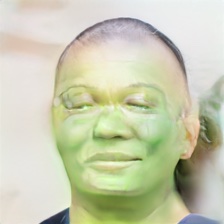} & \includegraphics[width=.16\linewidth,valign=m]{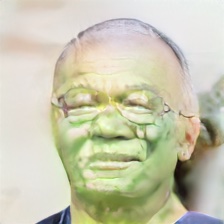}\vspace{0.2cm} \\
\makecell{Full} & \includegraphics[width=.16\linewidth,valign=m]{ablation/02492.jpg} & \includegraphics[width=.16\linewidth,valign=m]{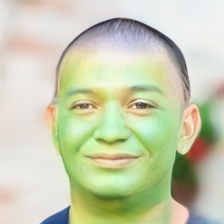} & \includegraphics[width=.16\linewidth,valign=m]{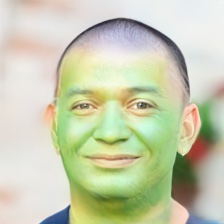} & \includegraphics[width=.16\linewidth,valign=m]{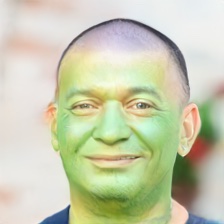} & \includegraphics[width=.16\linewidth,valign=m]{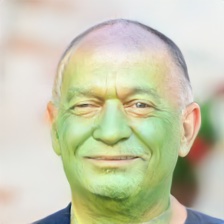}\vspace{0.2cm} \\
    
    \end{tabular}
    
    \caption{Ablation study: impact of the skip connections (SC.) and the style encoder}
    \label{tab:ablation_table4}
\end{figure}

Figure \ref{tab:ablation_table4} introduces the corresponding images for the qualitative ablation study made in our main work for the architecture design. It can be observed that the final architecture (\textit{Full}) retains identity, details, and gender better than the alternatives, even in challenging examples such as the second image.

\begin{figure}[t]
    \centering
    \begin{tabular}{llllll}
    \multicolumn{1}{c}{} & \multicolumn{1}{c}{Original} & \multicolumn{1}{c}{25} & \multicolumn{1}{c}{35} & \multicolumn{1}{c}{45} & \multicolumn{1}{c}{60} \\ 
    \makecell{Top-class GB} & \includegraphics[width=.16\linewidth,valign=m]{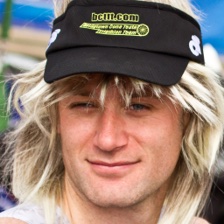} & \includegraphics[width=.16\linewidth,valign=m]{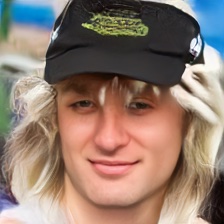} & \includegraphics[width=.16\linewidth,valign=m]{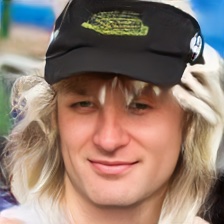} & \includegraphics[width=.16\linewidth,valign=m]{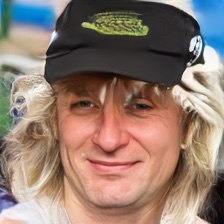} & \includegraphics[width=.16\linewidth,valign=m]{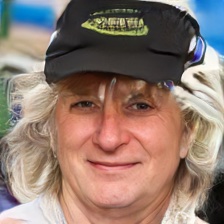}\vspace{0.2cm} \\
\makecell{Class-indep. \\ (Ours)} & \includegraphics[width=.16\linewidth,valign=m]{ablation/00907.jpg} & \includegraphics[width=.16\linewidth,valign=m]{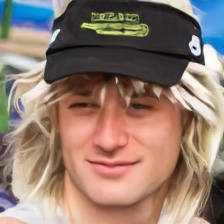} & \includegraphics[width=.16\linewidth,valign=m]{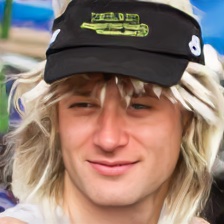} & \includegraphics[width=.16\linewidth,valign=m]{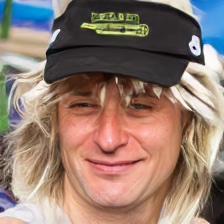} & \includegraphics[width=.16\linewidth,valign=m]{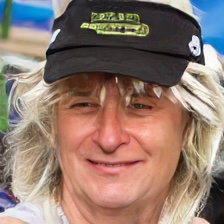}\vspace{0.2cm} \\
    
    \end{tabular}
    
    \begin{tabular}{llllll}
    \multicolumn{1}{c}{} & \multicolumn{1}{c}{Original} & \multicolumn{1}{c}{25} & \multicolumn{1}{c}{35} & \multicolumn{1}{c}{45} & \multicolumn{1}{c}{60} \\ 
    \makecell{Top-class GB} & \includegraphics[width=.16\linewidth,valign=m]{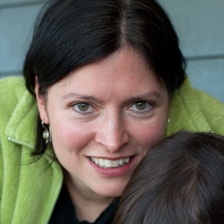} & \includegraphics[width=.16\linewidth,valign=m]{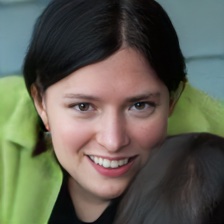} & \includegraphics[width=.16\linewidth,valign=m]{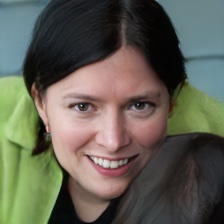} & \includegraphics[width=.16\linewidth,valign=m]{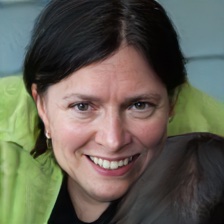} & \includegraphics[width=.16\linewidth,valign=m]{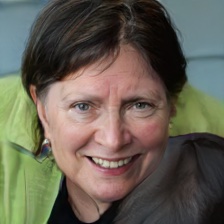}\vspace{0.2cm} \\
\makecell{Class-indep. \\ (Ours)} & \includegraphics[width=.16\linewidth,valign=m]{ablation/10599.jpg} & \includegraphics[width=.16\linewidth,valign=m]{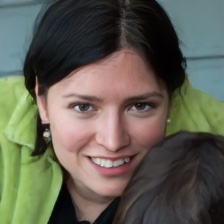} & \includegraphics[width=.16\linewidth,valign=m]{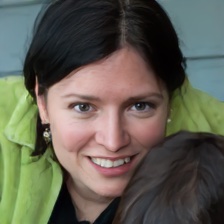} & \includegraphics[width=.16\linewidth,valign=m]{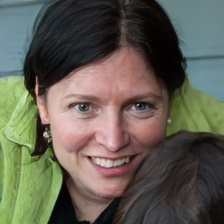} & \includegraphics[width=.16\linewidth,valign=m]{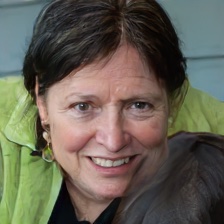}\vspace{0.2cm} \\
    
    \end{tabular}
    
    \begin{tabular}{llllll}
    \multicolumn{1}{c}{} & \multicolumn{1}{c}{Original} & \multicolumn{1}{c}{25} & \multicolumn{1}{c}{35} & \multicolumn{1}{c}{45} & \multicolumn{1}{c}{60} \\ 
    \makecell{Top-class GB} & \includegraphics[width=.16\linewidth,valign=m]{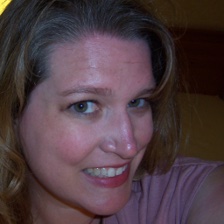} & \includegraphics[width=.16\linewidth,valign=m]{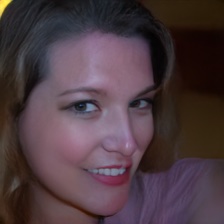} & \includegraphics[width=.16\linewidth,valign=m]{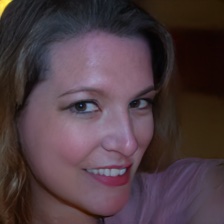} & \includegraphics[width=.16\linewidth,valign=m]{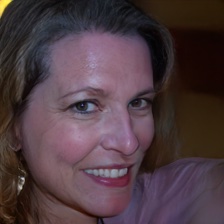} & \includegraphics[width=.16\linewidth,valign=m]{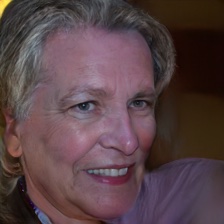}\vspace{0.2cm} \\
\makecell{Class-indep. \\ (Ours)} & \includegraphics[width=.16\linewidth,valign=m]{ablation/17639.jpg} & \includegraphics[width=.16\linewidth,valign=m]{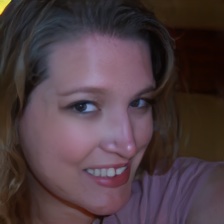} & \includegraphics[width=.16\linewidth,valign=m]{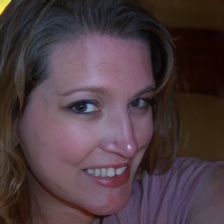} & \includegraphics[width=.16\linewidth,valign=m]{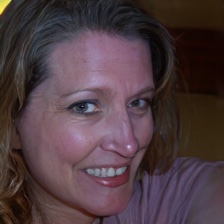} & \includegraphics[width=.16\linewidth,valign=m]{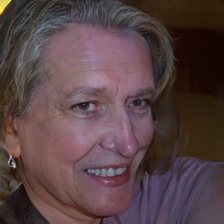}\vspace{0.2cm} \\
    
    \end{tabular}

    \caption{Ablation study: impact of the masking strategy used in CUSP.}
    \label{tab:ablation_table5}
\end{figure}

On the other hand, in Figure \ref{tab:ablation_table5}, the images corresponding to the masking strategy ablation are shown. Even though the differences are subtle,  the class-independent approach presents fewer artifacts while preserving details and identity better.

\subsection{Additional quantitative results}
In this section, we quantitatively study the impact of the blurring approach. We compare four approaches: the three masking configurations employed in the experiments of the main paper (HP, CP, and LP) and a \emph{Global blur} approach that uniformly blurs all images during training. This \emph{Global blurring} approach is equivalent to setting $\sigma_m$ to $0$ and $\sigma_g$ to $9$. Qualitative results are reported in Table \ref{tab:ablation_blur}.
\begin{table}[]
\centering
\caption{Blurring approach ablation on the \textit{FFHQ-RR} test set. CUSP HP (High preservation), CP (Custom preservation), and LP (Low preservation) are run with $(\sigma_m,\sigma_g)=(0,0)$, $(\sigma_m,\sigma_g)=(8,1.8)$ and $(\sigma_m,\sigma_g)=(8,4.5)$ respectively.\textit{ Global blur} is trained and run with $(\sigma_m,\sigma_g)=(0,9)$.}
\label{tab:ablation_blur}

\begin{tabular}{lrrr}
\toprule

                     &
                    \multicolumn{1}{c}{\textit{Reconstruction}} & \multicolumn{2}{c}{\textit{Age translation}}                          \\
                                & \multicolumn{1}{l}{\bf LPIPS ($\bf \times 10$)}             & \multicolumn{1}{l}{\textbf{Age MAE}} & \multicolumn{1}{l}{\bf Mean FID} \\ \midrule

Global blur              & 1.56                                         &  \bf 5.84                                  & 109.03                       \\
CUSP - LP & 1.09 &             6.07    &   \bf 104.44  \\ 
CUSP - CP &  0.78 &           6.29     &    104.63   \\
CUSP - HP & \bf 0.71                                         & 9.05                                  & 106.78                       \\ 
\bottomrule
\end{tabular}%

\end{table}

\begin{figure*}\centering
\includegraphics[width=0.7\textwidth]{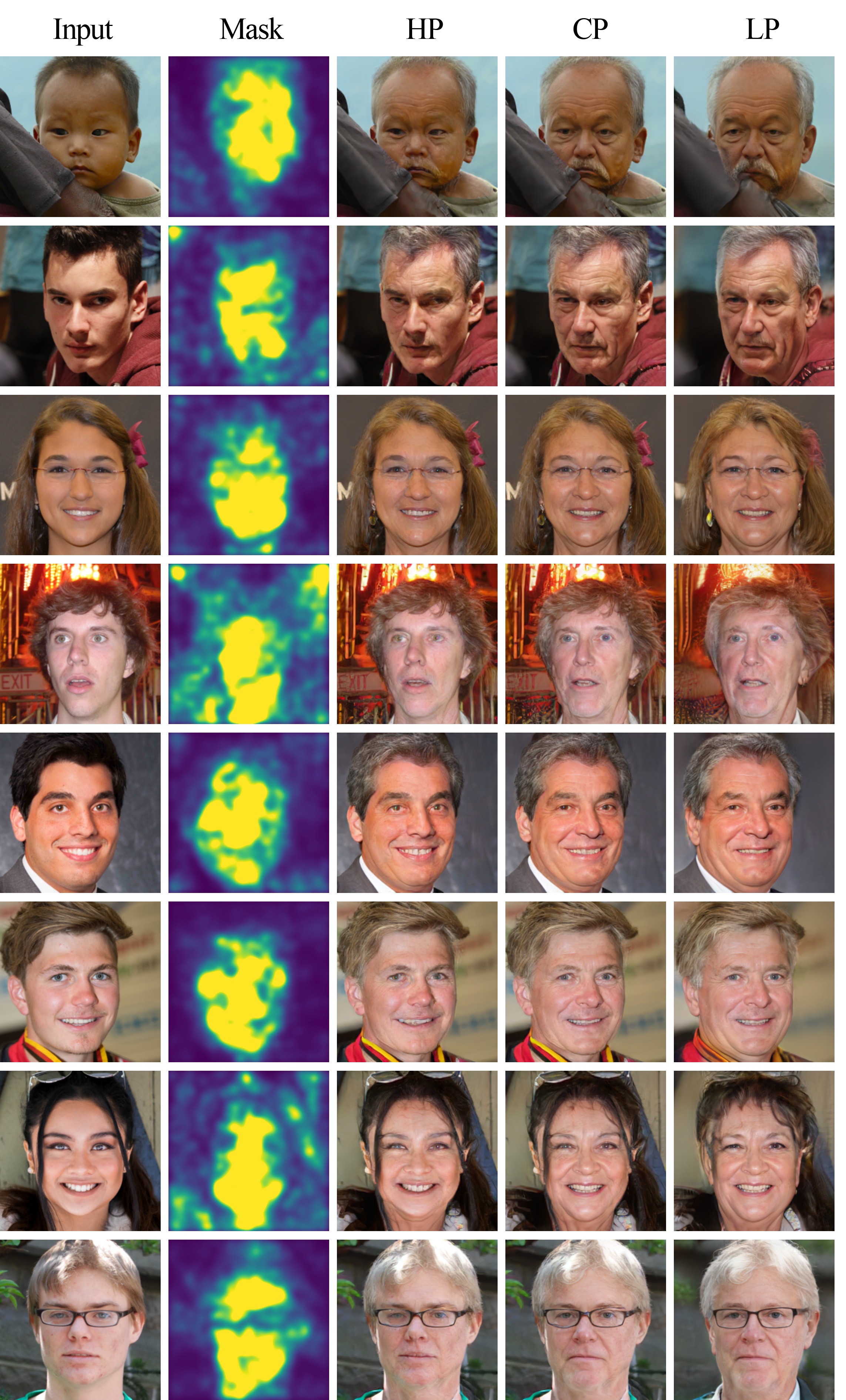}
  \caption{Qualitative study of the impact of the kernel value in CUSP at test time on young (lower than 30) to old (60). We compare images obtained with High, Low, and Custom structure preservation (referred to as LP, HP, and CP).   HP:$(\sigma_m,\sigma_g)=(0,0)$; CP:$(\sigma_m,\sigma_g)=(9,0)$; HP:$(\sigma_m,\sigma_g)=(9,9)$. The second column shows the mask estimated by our CUSP module with a color scale from blue (for 0) to yellow (for 1).}
  \label{fig:blur_ablation_big_y2o}
\end{figure*}

It shows that the two models with lower structure preservation (\emph{Global blur} and LP) obtain the best age translation scores, while their high reconstruction error (\ie LPIPS) shows that the details of the images are not preserved. On the contrary, CUSP-HP achieves better reconstruction at the cost of worse age translation scores. Our CUSP model with Custom blur parameters leads to a satisfying trade-off between reconstruction and age translation.

These experiments again justify the usefulness of letting the user the possibility to choose its own trade-off at inference time, as well as the use of different values for $\sigma_m$ and $\sigma_g$.

\subsection{Additional qualitative results}
\begin{figure*}[t]\centering
\includegraphics[width=0.7\textwidth]{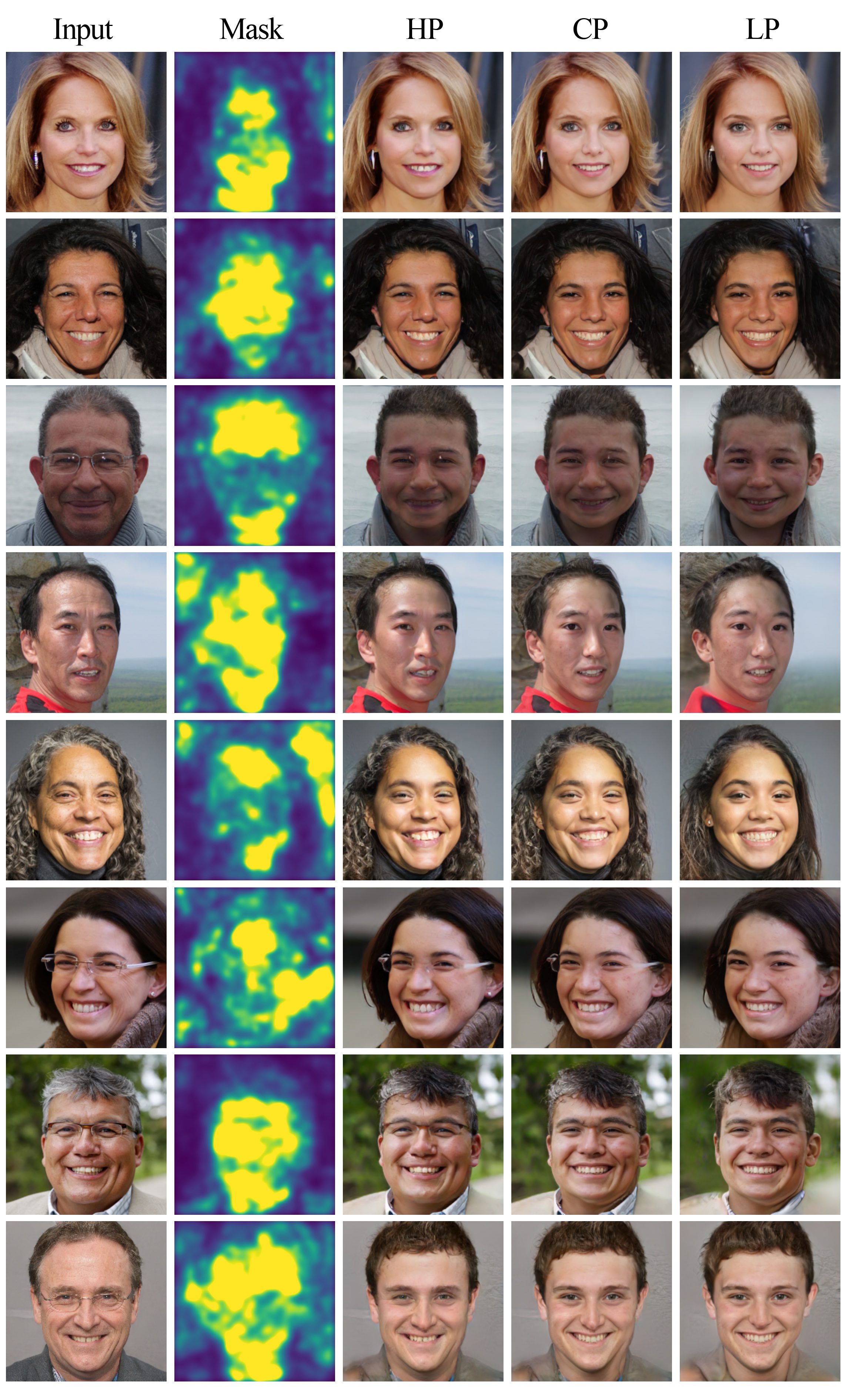}
  \caption{Qualitative study of the impact of the kernel value in CUSP at test time on old (between 50 and 60) to young (25). We compare images obtained with High, Low, and Custom structure preservation (referred to as LP, HP, and CP).   HP:$(\sigma_m,\sigma_g)=(0,0)$; CP:$(\sigma_m,\sigma_g)=(9,0)$; HP:$(\sigma_m,\sigma_g)=(9,9)$. The second column shows the mask estimated by our CUSP module with a color scale from blue (for 0) to yellow (for 1).}
  \label{fig:blur_ablation_big_o2y}
\end{figure*}

We now qualitatively evaluate the three masking configurations employed in the experiments of the main paper (HP, CP, and LP). 
Results on the \textit{FFHQ-RR} dataset are shown in  Figs. \ref{fig:blur_ablation_big_o2y} and \ref{fig:blur_ablation_big_y2o} for two different settings, old to young and young to old respectively.
Similar to the main paper, these results show that the high structure preservation variant preserves the face shape and hair growth. Meanwhile, the lower structure preservation allows stronger modifications of the face. We can see that the intermediate model with custom preservation achieves a satisfying trade-off.

\begin{figure*}[t]\centering
\includegraphics[width=0.7\textwidth]{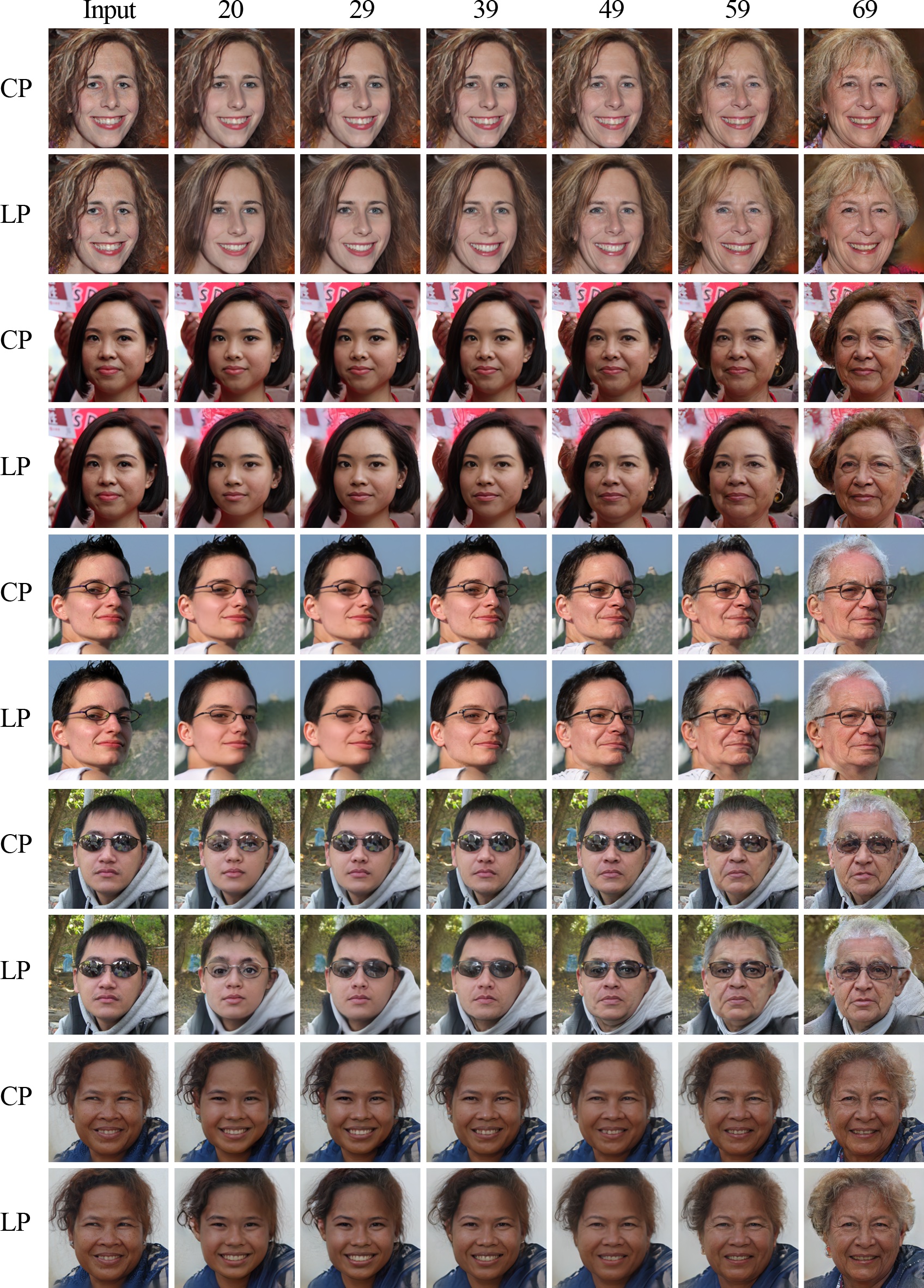}
  \caption{ Qualitative study of the impact of kernel values in CUSP at test time on the \textit{FFHQ-RR} test set. For each rows pair, the first corresponds to Custom structure preservation or CP $(\sigma_m,\sigma_g)=(0.9,7.2)$, the second to Low preservation, LP $(\sigma_m,\sigma_g)=(8.6,7.2)$.} 
  \label{fig:progresion_ablat_1}
\end{figure*}

\begin{figure*}[t]\centering
\includegraphics[width=0.7\textwidth]{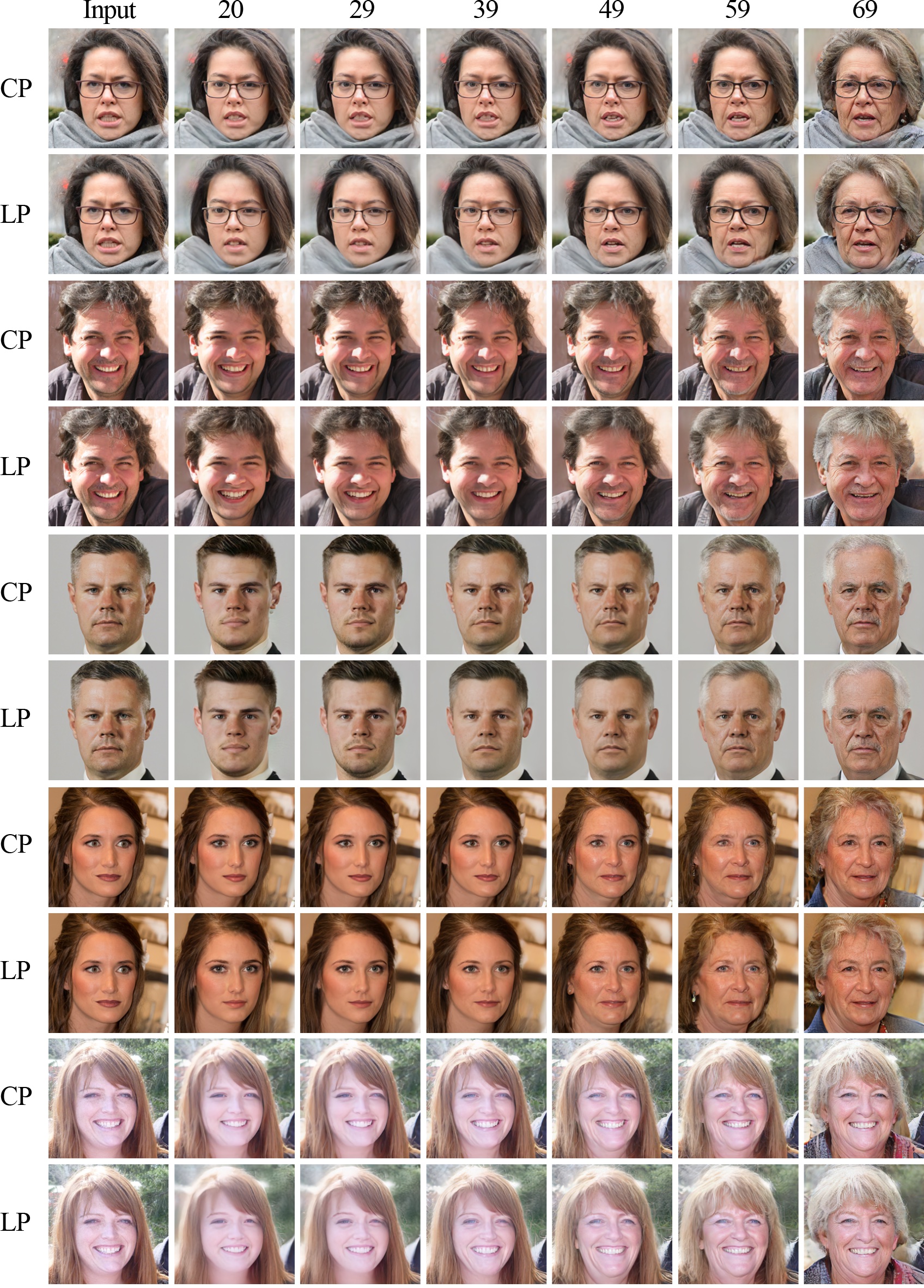}
  \caption{ Qualitative study of the impact of the kernel value in CUSP at test time on the \textit{FFHQ-RR} test set. For each rows pair, the first corresponds to Custom structure preservation $(\sigma_m,\sigma_g)=(0.9,7.2)$, the second to Low preservation $(\sigma_m,\sigma_g)=(8.6,7.2)$.}
  \label{fig:progresion_ablat_2}
\end{figure*}

Finally, both CP and LP are evaluated on six different target ages and two CUSP configurations in Figs. \ref{fig:progresion_ablat_1} and \ref{fig:progresion_ablat_2} on the  \textit{FFHQ-RR} dataset. Again, we see that lower preservation (\ie LP) applies strong modifications to the images to match the target age at the cost of lower preservation of identity (\eg see hairs in the first row of \ref{fig:progresion_ablat_2})

\subsection{Failure cases}

\begin{figure}[t]\centering
\includegraphics[width=0.8\columnwidth]{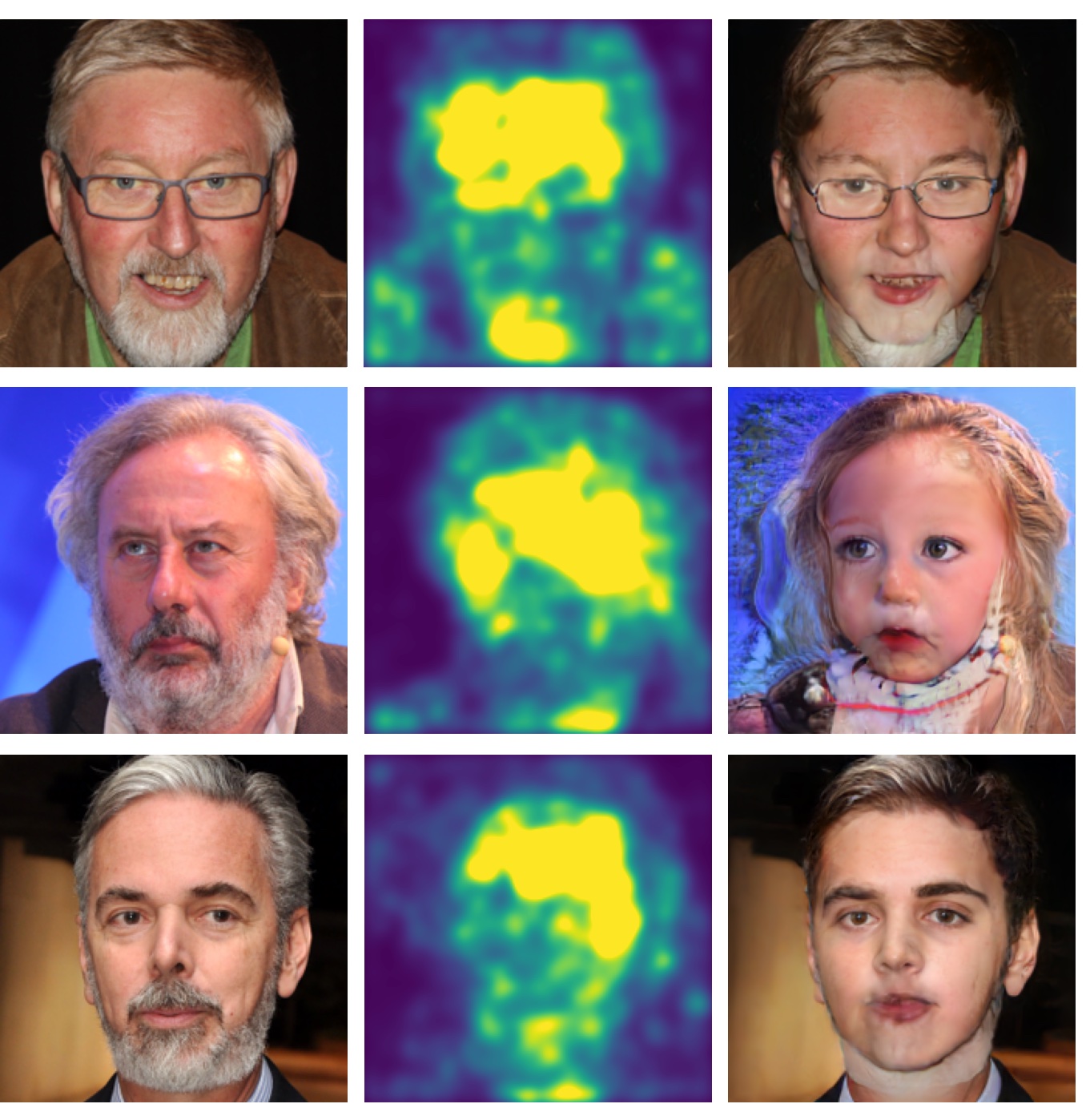}
  \caption{ From left to right: Input image, $\Mmat$ Mask, and target age 25. The resulting images are obtained with the CUSP CP setting $(\sigma_m,\sigma_g)=(8,1.8)$ on \textit{FFHQ-RR}.}
  \label{fig:old_failures}
\end{figure}
In Fig. \ref{fig:old_failures}, we show some failure cases.
These images present noticeable artifacts. We observe that these artifacts appear mainly in the presence of white beards. The CUSP mask does not entirely select beards (see the second row). This behavior might indicate an inherited bias from IMDb-Wiki, where white beards can be rare in celebrities' pictures.

In these images, we also see some saturation artifacts similar to those described in \cite{stylegan2}. Even though we employ \textit{Weight demodulation} \cite{stylegan2} that is aimed at solving this issue, we observe that some artifacts remain.

\section{Comparison with State-of-the-Art}
\label{sec:sota}
We now report additional qualitative comparison with State-of-the-Art. We report separate comparisons with HRFAE and LATS as in the main paper.
\subsection{Comparison with HRFAE}

In Fig.~\ref{fig:hrfae_comp}, we show some additional comparisons with HRFAE on the \textit{FFHQ-RR} dataset using the CUSP CP setting. These results are in line with the results reported in the main paper. In addition, we observe that our approach is able to apply more substantial modifications to the image to better match the target age.
\begin{figure*}[t]\centering
\includegraphics[width=0.7\textwidth]{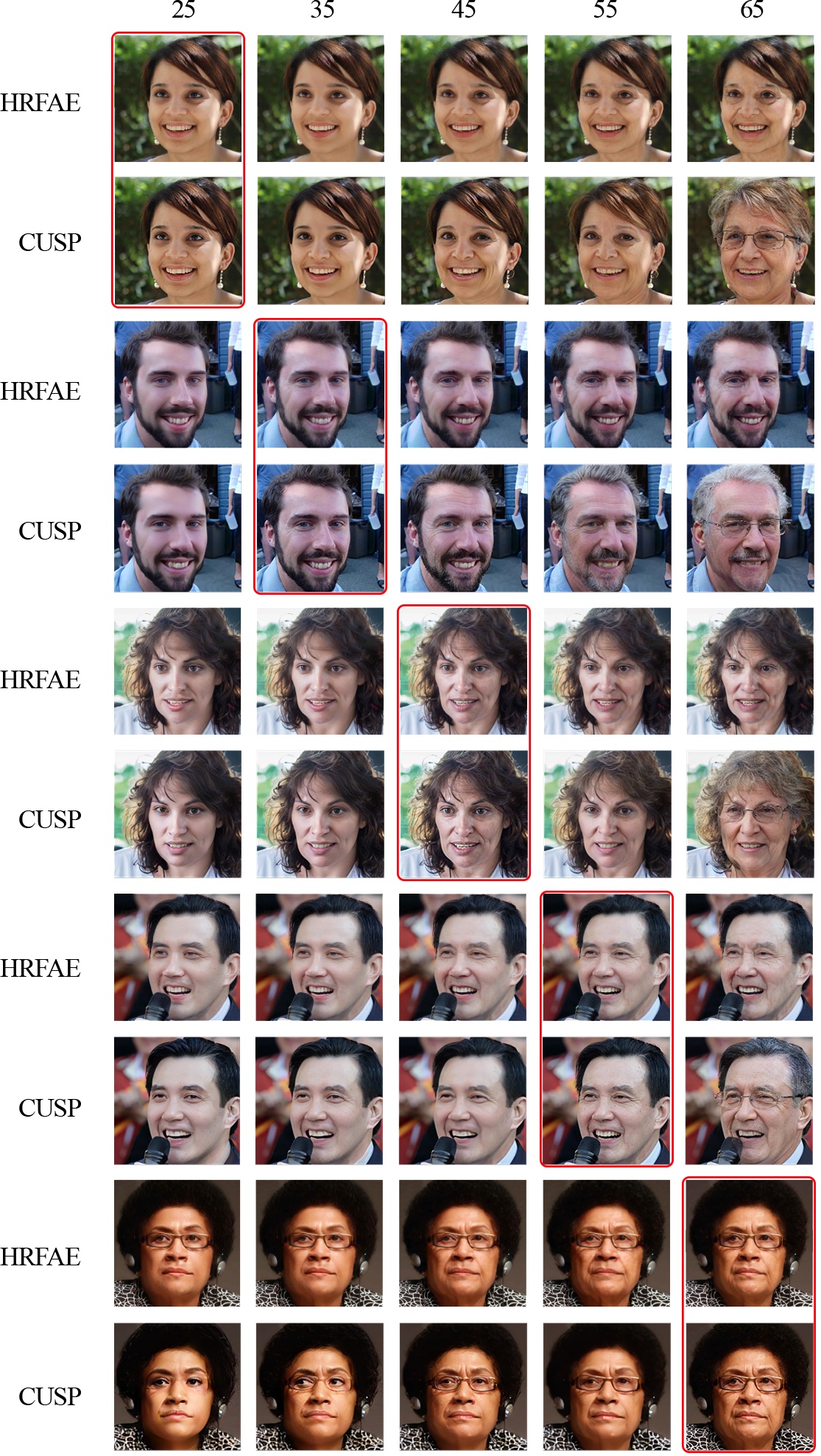}
  \caption{ Qualitative comparison with HRFAE for different age targets on \textit{FFHQ-RR}. The used setting is CUSP CP $(\sigma_m,\sigma_g)=(8,1.8)$. The images corresponding to the target ages are highlighted with red frames.}
  \label{fig:hrfae_comp}
\end{figure*}

We complete this comparison by showing the results obtained with our method on the same examples previously used in \cite{hrfae2021} and using their qualitative evaluation. We generate images varying the target age from 20 to 69. Our results are smooth, and we observe that our proposed method is able to apply more profound changes in the case of extreme ages.

\begin{figure*}[t]\centering\centering
\includegraphics[width=\textwidth]{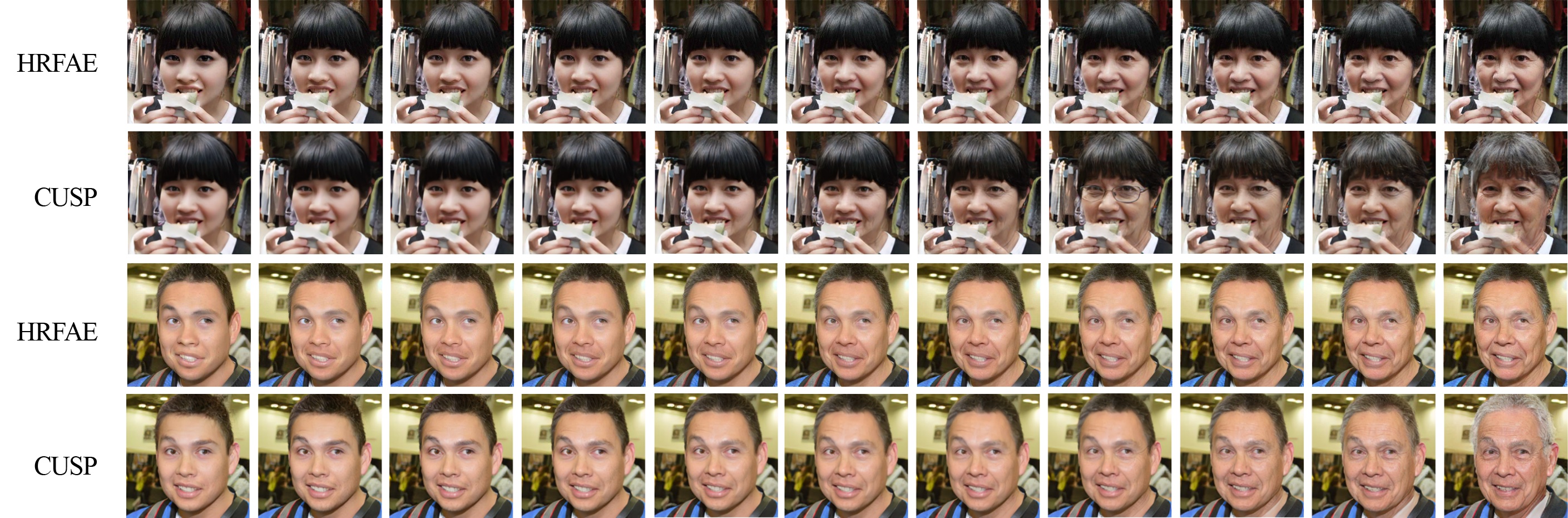}
  \caption{HRFAE comparison on \textit{FFHQ-RR} for smooth progression. The target age goes uniformly from 20 to 69.}
  \label{fig:hrfae_prog}
\end{figure*}

\subsection{Comparison with LATS}

\begin{figure*}[t]\centering\centering
\includegraphics[width=0.7\textwidth]{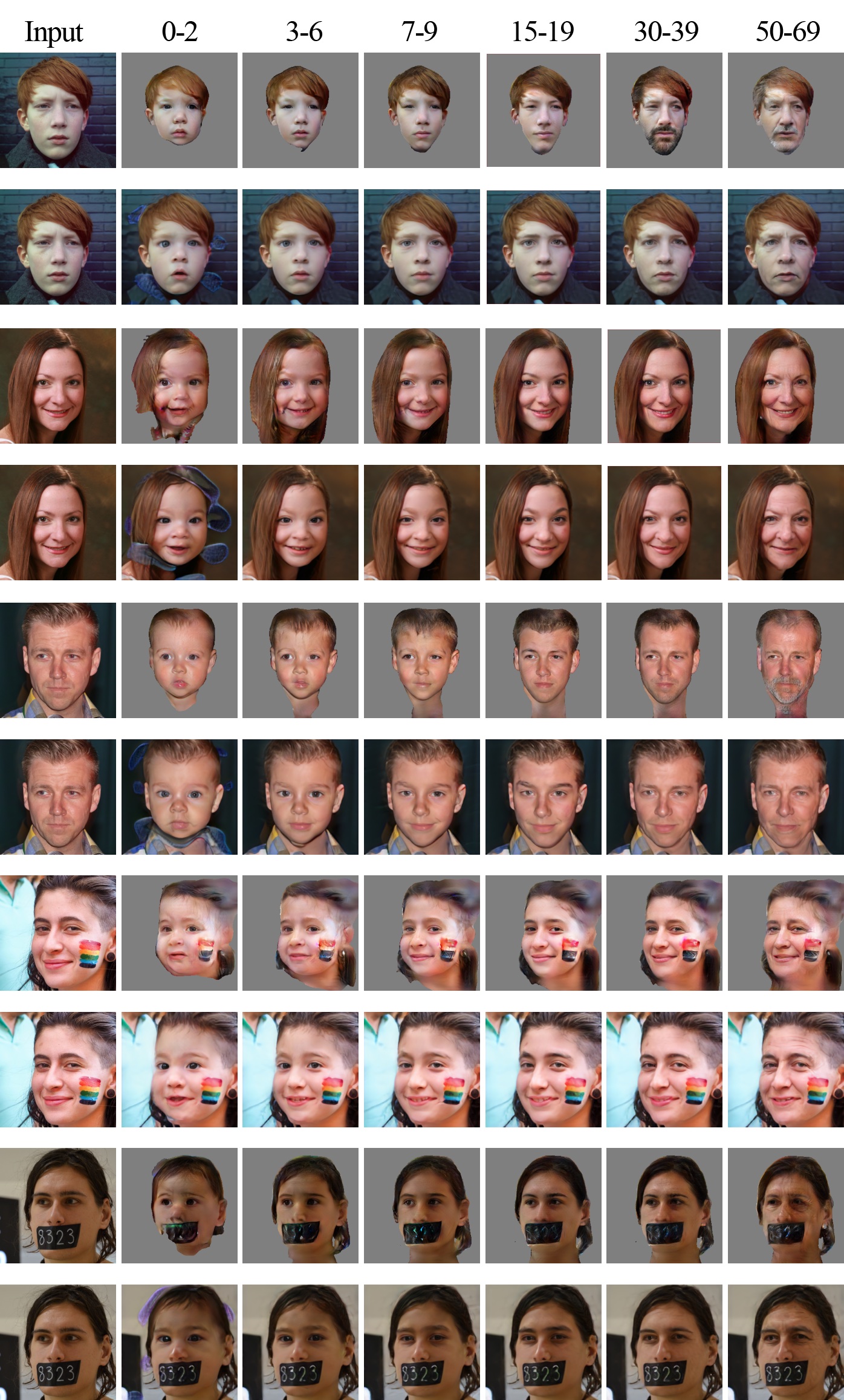}
  \caption{ Qualitative comparison with LATS on \textit{FFHQ-LS} test set for different age targets.}
  \label{fig:lats_comp_big}
\end{figure*}

In Fig.~\ref{fig:lats_comp_big}, we show a complementary qualitative comparison with LATS on several images. Similar to the results in the main paper, we observe that our method is on par with LATS while having the numerous advantages detailed in the main paper. 

\new{
  \subsection*{Pending comparison}  
Posteriorly to the submission of this work, \cite{alaluf2021matter} was published. The authors propose a new method for age editing built upon a collection of pretrained networks and custom-trained modules (methodology described in our main paper). Even though they inherit the same flaws pretrained StyleGAN2 and pSp \cite{richardson21cvpr} have (\ie blurry backgrounds and low structural and identity preservation), they achieve promising deep structural age-related transformations. For this, comparing both methods in the future should be interesting.}

\section{Datasets licenses}
\label{sec:licence}
Both CelebA-HQ and FFHQ are publicly available and widely used datasets. CelebA-HQ is openly available for its use in research but has some rights reserved. FFHQ is made available under the Creative Commons BY-NC-SA 4.0 license. Thus every derivative (e.i., FFHQ-RR and FFHQ-LS) have the same license.

\end{document}